\documentclass{article} %
\usepackage{iclr2027_conference,times}

\usepackage{amsmath,amsfonts,bm}

\def\eqref#1{equation~\ref{#1}}

\def\1{\bm{1}}

\DeclareMathAlphabet{\mathsfit}{\encodingdefault}{\sfdefault}{m}{sl}
\SetMathAlphabet{\mathsfit}{bold}{\encodingdefault}{\sfdefault}{bx}{n}

\usepackage{hyperref}
\usepackage{url}
\usepackage{graphicx}
\usepackage{enumitem}
\usepackage{multirow}
\usepackage{booktabs}
\usepackage[most]{tcolorbox}
\usepackage{xcolor}
\usepackage{fancyvrb}
\usepackage{fvextra}
\usepackage{tabularx}
\usepackage{amssymb}
\newtcolorbox{llmprompt}[1]{
    breakable,
    colback=blue!5!white,    %
    colframe=blue!75!black,  %
    coltitle=white,          %
    fonttitle=\bfseries,     %
    fontupper=\ttfamily,     %
    title=#1,                %
    arc=4mm,                 %
    boxrule=1pt,             %
    enhanced
}

\title{Ground, Iterate, Trace: Decomposing AI \\ Research Ideation}

\author{Samson Yu$^{1}$, Qiran Zou$^{1}$, Jing Yu Lim$^{1}$, Rushi Shah$^{1}$, Anirudh Goyal$^{2}$, Dianbo Liu$^{1}$ \\
{\normalfont $^{1}$National University of Singapore \quad $^{2}$Meta Superintelligence Labs}
}

\iclrfinalcopy %

\begin{document}

\maketitle
\lhead{Preprint. Under review.}

\begin{abstract}
Large language models have been widely used to generate plausible research ideas, yet it remains unclear what makes automated ideation effective and how generated novel proposals arise from prior scientific work. We introduce the \textbf{Creative AI Researcher (CAR)}, an agentic framework combining literature grounding, iterative development and explicit conceptual operators. CAR constructs a semantic knowledge graph from retrieved literature and equips an ideation agent with graph-retrieval tools and four operators: bisociation, analogy, exploration and elaboration. Each operator call is checked against structural acceptance conditions and recorded, producing a trace of intermediate transformations and declared literature sources. We evaluate CAR with two language-model backbones across 40 LiveIdeaBench research topics and two execution benchmarks. CAR achieves significantly higher methodological novelty (MN) than single-shot prompting and The AI Scientist. Its methodological novelty does not differ significantly from the Research Ideation Agent, while its feasibility scores are higher on both backbones. With a shared experiment executor, CAR achieves the highest pairwise win rate among the evaluated AI methods on FML-bench-Lite and BioXArena. In a ten-condition ablation study, MN point estimates favor CAR over one-step proposal on both backbones, but no MN contrast survives correction for multiple comparisons. The results do not establish a consistent quality benefit from explicit operators beyond free-form iteration. CAR makes the sequence of operations and declared literature sources inspectable, separating judged idea quality from procedural traceability.
\end{abstract}

\section{Introduction}
\label{introduction}
AI scientists, LLM agents that propose research ideas, implement them and report the results, have moved from demonstrations to systems whose papers are submitted for peer review \citep{lu2026towards}. In these systems, ideation is either a brief step before execution or is interleaved with it, as the agent proposes its next change to a working implementation. How the idea is formed matters for what follows: given the same executor, replacing its own ideas with ideas formed separately changes the outcome (\S\ref{results_effectiveness}). Furthermore, the diversity of an agent's ideas predicts its
research performance on MLE-bench \citep{audran2025does}.

Current ideation agents fall short in two opposite ways. Agents prompted to be bold produce ideas that experts judge more novel than human ideas but slightly weaker on feasibility \citep{si2025can}, and the same agent is the least feasible of those we compare (\S\ref{results_originality}). The ideation stages of end-to-end AI scientists sit at the other end: run on their own, they produce the most feasible ideas we measure and, for The AI Scientist on GPT-5.4, the least original. These contrasting profiles motivate evaluating originality and effectiveness together, the two criteria of the standard definition of creativity \citep{runco2012standard}.

We address the step before implementation: forming a research proposal by transforming the concepts and relations found in prior work. Theories of creativity describe such transformations, among them bisociation \citep{koestler1964act}, analogy \citep{gentner1983structure}, exploration of a conceptual space \citep{boden2004creative} and elaboration \citep{guilford1967nature}. In LLM ideation, such processes are often elicited through prompting \citep{gu2024llms,shen2026unlocking}, while some systems
also retain structured intermediate representations.
This motivates distinguishing the instruction to make a
creative move from an explicit check of its form. Ideation agents also bundle retrieval and multi-round refinement into one pipeline \citep{si2025can,li2025chain,hu2024nova}, making their individual contributions difficult to disentangle. This leaves two questions open: which parts of an ideation process improve its ideas, and how a given idea arose from the literature.

We introduce the \textbf{Creative AI Researcher (CAR)}. CAR extracts a semantic knowledge graph of problems, methods, results, claims and limitations from the literature retrieved for a topic. An ideation agent then develops a proposal over several rounds, using graph-retrieval tools and the four processes above, which CAR exposes as \emph{operators} that edit the graph. Each operator has a structural definition that is validated before the edit is applied. The accepted and rejected operations, and the literature entities the final proposal declares, form a trace for every idea.

CAR allows us to vary iterative development, operator
availability, validation and relation labels within one
implementation. We evaluate it on $40$ LiveIdeaBench topics
with two LLM backbones and execute its ideas with a shared
experiment agent on FML-bench-Lite and BioXArena.
CAR has higher methodological novelty (MN) than single-shot
prompting and The AI Scientist. Its MN differences from the
Research Ideation Agent (RIA) \citep{si2025can} are not
significant, while its feasibility scores are higher on
both backbones.
CAR also obtains the highest pairwise win rate among AI
methods on both execution benchmarks.
The ablation MN point estimates favor CAR over one-step proposal,
but do not establish a consistent quality benefit from explicit
operators beyond free-form iteration.
The operators provide a recorded sequence of graph edits
and declared literature sources.

Our contributions are:
\begin{itemize}[leftmargin=*,noitemsep]
  \item A decomposition of AI research ideation into literature
    grounding, iterative development and explicit creative operators. A ten-condition study on two backbones examines
    iteration and the operator layer within this grounded setting,
    distinguishing their effects on judged idea quality from
    the procedural trace they provide.
  \item CAR, which makes the decomposition testable: four creativity processes implemented as validated, recorded operators over a literature-derived knowledge graph, so that every proposal carries a trace from prior work to the final idea.
  \item An evaluation of CAR on originality and effectiveness, the two criteria of the standard definition of creativity, on three benchmarks, with judge agreement assessed against expert ratings.
\end{itemize}

\section{Related work}
\label{related_work}

\subsection{Grounding ideation in the literature}
\label{rw_ground}
Research ideation systems ground ideas in prior work \citep{wang2024scimon,yang2025moose,si2025can,baek2025researchagent} or through a graph, which selects or forecasts concept pairs \citep{gu2024generation,gu2025forecasting}, conditions generation \citep{li2026graph2idea,lei2025motivgraph}, supplies reasoning paths \citep{buehler2024accelerating,ghafarollahi2025sciagents}, is extended by the model as it reasons \citep{buehler2025situ,buehler2025agentic}, or holds the ideas themselves, as in Evolving Idea Graphs \citep{dong2026evolving}. Evolving Idea Graphs also validates candidate graph edits. CAR focuses on four creative operators with explicit acceptance conditions, recording their accepted and rejected calls as ideas develop.

\subsection{Iterative and agentic ideation}
\label{rw_iterate}
Chain of Ideas \citep{li2025chain} and Nova \citep{hu2024nova} retrieve and refine seed ideas over several rounds, multi-agent teams assign roles and debate \citep{baek2025researchagent,ghafarollahi2025sciagents,swanson2025virtual,gottweis2025towards,su2025many}, and end-to-end AI scientists couple ideation to experiment search and writing \citep{lu2026towards,tang2026ai,weng2026deepscientist}, with evidence from MLE-bench that ideation diversity predicts performance \citep{audran2025does}. Many of these systems develop ideas through natural-language revision and critique before or during experimental execution. CAR develops proposals before execution, combining iterative
reasoning with explicit creative operators whose graph edits
are validated and recorded.

\subsection{Creative operators and traceable reasoning}
\label{rw_trace}
CAR's operators are named after bisociation \citep{koestler1964act}, with a bridge entity from conceptual blending \citep{fauconnier2003conceptual}, structure-mapping analogy \citep{gentner1983structure}, exploratory creativity \citep{boden2004creative,wiggins2006preliminary} and elaboration \citep{guilford1967nature}. Earlier computational work includes analogy mining \citep{hope2017accelerating} and bisociative networks \citep{ahmed2018creative}, CHIMERA catalogs recombinations in published papers \citep{sternlicht2025chimera}, and LLM work elicits such moves through prompting \citep{radensky2026scideator,gu2024llms,shen2026unlocking} or implements combinational, exploratory and transformative
reasoning over generated thoughts \citep{suzuki2025universe}.
Some approaches retain intermediate representations:
\citet{radensky2026scideator} record purpose, mechanism and evaluation facets, while \citet{shen2026unlocking} generate and assess explicit analogies.
Such artifacts make the proposed transformations inspectable,
but do not establish that the final idea depends on them.
This distinction also arises in the faithfulness problem of
chain-of-thought \citep{turpin2023language,lanham2023measuring}.
\citet{lyu2023faithful} make reasoning faithful by having an LLM
translate a problem into a symbolic representation, which a deterministic solver executes to obtain the answer.
CAR applies a related principle at the tool boundary: each creative operator must satisfy structural acceptance conditions before its graph edit is applied. The resulting trace records the operations performed,
without establishing their causal necessity to the final proposal.

\subsection{Evaluating AI-generated research ideas}
\label{rw_eval}
Benchmarks pair LLM judges with fixed rubrics \citep{ruan2026evaluating,qiu2025ai} or test whether judges predict later research outcomes \citep{ye2026proof}. Retrieval-grounded checks compare an idea against its nearest published work \citep{si2025can,shen2026unlocking,liu2026harnessing,shahid2025literature}. Whether LLMs are qualified originality reviewers is contested
\citep{huang2025large}.
We assess agreement between our originality and methodological
novelty judges and the expert ratings of \citet{si2025can} and an adjusted
human-agreement reference, with development-set limitations
discussed in Appendix \ref{app:validation}.

\section{CAR: Grounded creative iteration over a knowledge graph}
\label{method}

\begin{figure}[!t]
\centering
\includegraphics[width=\linewidth]{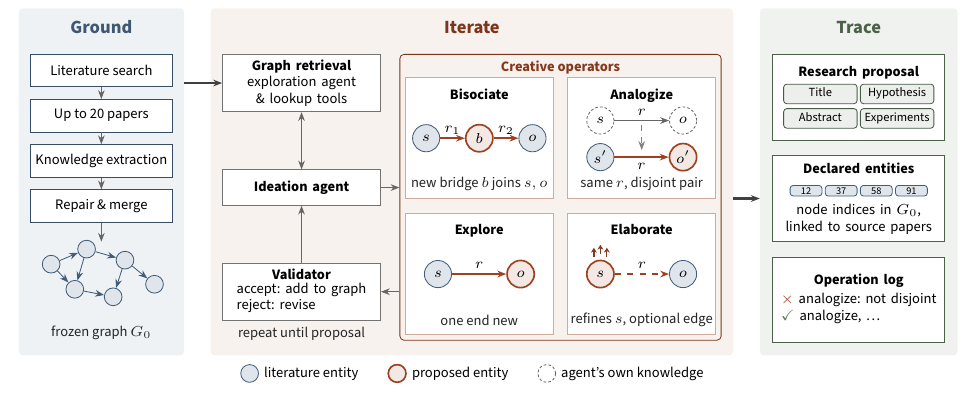}
\caption{Overview of CAR. \textbf{Ground:} a typed knowledge graph is extracted from the retrieved literature, repaired and frozen. \textbf{Iterate:} the ideation agent develops an idea with four creative operators, each validated against the graph and applied only when it is successful. \textbf{Trace:} the proposal is stored with its declared literature entities and its accepted and rejected operations.}
\label{fig:overview}
\end{figure}

CAR has three components (Figure \ref{fig:overview}). Given a research aim, it first builds a \textit{semantic knowledge graph} (KG) of scientific entities and relations from retrieved literature (\S\ref{knowledge_construction}). An LLM ideation agent then develops ideas over this graph with graph-retrieval tools (\S\ref{graph_retrieval}) and four validated \textit{creative operators} (\S\ref{research_ideation}). Each idea is stored with a \textit{trace} of the operations and declared literature entities.

\subsection{Knowledge construction}
\label{knowledge_construction}

\textbf{Corpus.} For each research task, an LLM queries the Semantic Scholar API in an iterative loop, with search prompts adapted from the Research Ideation Agent \citep{si2025can}: keyword queries, queries for papers similar to a given paper, and queries for a paper's references (Appendix \ref{app:prompts_knowledge_search}). An LLM scores the retrieved papers for relevance and we retain up to $20$ papers with the highest relevance scores among those whose full text can be obtained.

\textbf{Extraction.} From each paper an LLM extracts typed entities, each with a technical description, and the relations between them: the paper itself, and its problems, tasks, concepts, assumptions, methods, implementation details, hyperparameters, materials, experiments, metrics, results, claims, limitations and formulas (Appendix \ref{app:prompts_knowledge_extraction}). This follows a view from scientometrics that a paper's elements are the substrate of original recombination \citep{dirk1999measure, luo2022combination, zhao2025review}. Extracting them as distinct typed entities makes recombination expressible as an explicit operation on the graph. The result is a directed multigraph $G=(V,E)$ with $E \subseteq V \times R \times V$, where $R$ is the relation vocabulary produced by extraction. We write $V_{\mathrm{lit}}=V$ for these literature nodes and $V_{\mathrm{paper}}\subset V_{\mathrm{lit}}$ for paper nodes, which record provenance only. Two repair steps add relations for insufficiently connected entities and merge nodes with matching normalized names and entity types. Neither step introduces new entities (Appendix \ref{app:repair}). The graph is built once per topic and frozen.
Each idea starts from this graph, with relation labels permuted in the relation-label-shuffle ablation.

\subsection{Graph retrieval}
\label{graph_retrieval}
The ideation agent reads the graph through three tools. \texttt{ask\_knowledge\_graph\_exploration\allowbreak \_agent} delegates a natural-language question to a graph exploration agent adapted from ARK \citep{polonuer2026autonomous}, which searches the graph by keyword and over one-hop neighborhoods and returns relevant entities with its reasoning for each. \texttt{get\_entity\_details} returns an entity's full description given its index, and \texttt{get\_edges\_by\_entity} lists its incident literature edges.

\subsection{Research ideation with creative operators}
\label{research_ideation}

\textbf{Creative operators.} The ideation agent develops an idea through four creative operators, each an explicit edit to the graph that is \emph{validated before it is applied} and recorded afterwards. Let $G_t=(V_t,E_t)$ be the graph after $t$ accepted operations within one idea, with $G_0=G$, and let $V_{\mathrm{prop},t}=V_t\setminus V_{\mathrm{lit}}$ be the entities proposed so far. Every proposed edge $(s,r,o)$ must carry a written description of why $s$ and $o$ stand in relation $r$ (which need not belong to $R$), every new entity must carry a type and a technical description, and no operation may name a paper node: it must use the specific problem, method, result or claim. An operation that meets its condition in Table \ref{tab:operators} is applied. Otherwise, the agent receives an error naming the violated condition and must revise it or choose another move.

\begin{table}[!t]\centering\small
\caption{Creative operators and their acceptance conditions. Triplets $(s,r,o)$ are edges from subject $s$ to object $o$ under relation $r$. $u\sim_t v$: $E_t$ contains an edge between $u$ and $v$ in either direction. All named existing entities must lie outside $V_{\mathrm{paper}}$.}
\label{tab:operators}
\begin{tabular}{@{}lll@{}}
\toprule
Operator & Arguments & Acceptance conditions \\
\midrule
\textit{bisociate} & $(s,r_1,b),\,(b,r_2,o)$ & $s,o\in V_t,\ \ s\neq o,\ \ s\not\sim_t o,\ \ b\notin V_t$ \\
\textit{analogize} & source $(s,r,o)$, target $(s',r',o')$ & $r'=r,\ \ \{s',o'\}\cap\{s,o\}=\varnothing,\ \ \{s',o'\}\cap V_t\neq\varnothing$ \\
\textit{explore} & $(s,r,o)$ & $|\{s,o\}\cap V_t|=1$ \\
\textit{elaborate} & $s$, new description, optional $(s,r,o)$ & $s\in V_{\mathrm{prop},t},\ \ o\in V_{\mathrm{lit}}$ \\
\bottomrule
\end{tabular}
\end{table}

\textbf{Operator design.} \textit{Bisociate} \citep{koestler1964act} connects two existing, unlinked entities through a \emph{new} bridge entity. The one-edge form $(s,r,o)$ is not allowed, since it only asserts a relation between two concepts the literature already contains, whereas the bridge forces a new concept at their intersection \citep{fauconnier2003conceptual}. \textit{Analogize} follows structure mapping \citep{gentner1983structure}: it carries a relation from a source pair to a target pair, and only the target is added to the graph. The source may come from the agent's own knowledge, since restricting it to the corpus would confine transfer to the retrieved literature. The target must change both endpoints: keeping one fixed ($A\!:\!B \rightarrow C\!:\!B$) does not carry the relation to a new pair, it only adds $C$ as another entity related to $B$. \textit{Explore}, after Boden's exploratory creativity \citep{boden2004creative}, extends the graph by one step from an existing entity, which may itself be proposed. \textit{Elaborate}, after Guilford's elaboration \citep{guilford1967nature}, adds no entity: it replaces a proposed entity's description with a more specific one and may tie it to a literature entity through one grounding edge. Every operation therefore attaches to the graph at one or more entities, which is what makes an idea traceable.

\textbf{Proposal.} The agent is instructed to complete at least three accepted operations before submitting a proposal. The proposal follows the idea format of The AI Scientist-v2 \citep{lu2026towards} and additionally declares the literature entities it builds on as node indices in $V_{\mathrm{lit}}$.
We store each idea with its accepted and rejected operations, their arguments and their errors, which together with the declared entities form the \emph{trace} of the idea (Appendix \ref{app:traces}). Proposed entities and edges are discarded before the next idea, so every idea starts from $G_0$.

\section{Experiments}
\label{experiments}

We frame evaluation with the standard definition of creativity. We first assess (a) the \textit{originality} and \textit{feasibility} of CAR's ideas, and then go beyond feasibility, which is necessary but not sufficient for creativity, to assess (b) their \textit{effectiveness} as test-metric improvement when executed.

\subsection{Benchmarks and tasks}
\label{experiments_benchmarks}

\textbf{LiveIdeaBench} \citep{ruan2026evaluating} contains 1,180 scientific keywords from 22 domains and uses LLM judges to assess the ideas generated for a keyword along dimensions inspired by Guilford's account of divergent thinking \citep{guilford1967nature}: \textit{originality}, \textit{feasibility}, \textit{clarity} (how understandable and complete an idea is) and \textit{fluency} (how distinct a method's ideas for one keyword are from one another). We select 40 keywords, ten each from the computational, physical, life and social sciences. We report its fifth dimension, flexibility, in Appendix~\ref{app:liveideabench_full}. Our modifications to the protocol are listed in Appendix \ref{app:expsettings_liveideabench}.

\textbf{FML-bench} \citep{zou2026fml} comprises 18 machine learning research tasks in 10 domains, each with a baseline repository, a set of editable files and a held-out test metric, and scores an agent on how far it improves the baseline. All agents edit code and run experiments through the same benchmark-provided tools, so differences between agents reflect their research strategy rather than their tooling. We use the \textbf{Lite} split of eight tasks.

\textbf{BioXArena} \citep{li2026bioxarena} is a benchmark of 76 end-to-end biomedical machine learning tasks spanning nine areas, most combining several input modalities. Unlike FML-bench, a task supplies data and a metric but no baseline repository, so an agent must design the pipeline and the method. We use the ten tasks of the benchmark's human-comparison pilot, which supplies a human reference. One task, cell-type-from-expression, has a label leak that makes it trivially solvable (five of seven entries, including CAR and the human reference, score $1.000$). We show it but exclude it from means and win rates.

\subsection{Evaluation metrics}
\label{experiments_metrics}

\textbf{LiveIdeaBench.} We follow the benchmark's LLM-as-a-judge protocol, with its prompts, scales and fluency score map. A panel of three critics from vendors other than the generators' scores each idea $1$--$10$ for originality (O), feasibility (F) and clarity (C), and judges fluency (Flu.) on every pair of a method's five ideas for each keyword.

Because a critic judges originality from memory, we add a \emph{methodological novelty} (MN) judge that grounds the judgment in retrieved prior work. Each idea is converted to an approximately $250$-word abstract. Related papers are retrieved from Semantic Scholar with the iterative query loop of \citet{si2025can} under a publication-date cut-off, ranked by its LLM relevance score with ties broken by SPECTER2 similarity \citep{singh2023scirepeval}, and the top $10$ are kept. The same three judges then apply the Stratified Novelty Judge of \citet{shen2026unlocking}, adapted from biomedical to general research, rating $0$--$10$ how much of the proposed methodology has
already been applied to the idea's research task in the
retrieved papers, and we average the three ratings as the idea's overlap. We report $\mathrm{MN} = 10 - \text{overlap}$, and ideas with no eligible retrieved paper are excluded, not scored as maximally novel. Against the expert ratings of \citet{si2025can},
the originality critics and MN judge reach idea-level
Spearman $\rho=0.36$--$0.38$ and $0.39$, respectively.
The same sample informed MN evaluator development
(Appendix \ref{app:validation}).

We average scores within each keyword, then report the mean
and standard error across the $40$ keywords.
For each comparison, we subtract the comparator's mean score from CAR's within each keyword, using all scored ideas. The average difference therefore equals the difference between the reported overall means.

\textbf{FML-bench and BioXArena.} We report each benchmark's own test metric (Appendix \ref{app:expsettings}), rounded to its reported precision (three decimals) before any comparison, so that the extra precision of our own runs cannot break a tie the published baselines do not have. Pairwise win rate is the fraction of (task, comparator)
pairs in which an entry's rounded score is strictly higher,
so a tie is a win for neither (Appendix~\ref{app:ties}).

\subsection{Experimental settings}
\label{experiments_settings}

\textbf{LiveIdeaBench.} We compare CAR with a single-shot \textbf{Base} LLM (the backbone prompted once with the benchmark's own idea prompt), the \textbf{Research Ideation Agent} (RIA) \citep{si2025can}, and the ideation stages of \textbf{The AI Scientist} and \textbf{The AI Scientist-v2} \citep{lu2026towards}. Each baseline generates in its native format and is normalized afterwards (Appendix \ref{app:fidelity}). RIA draws its retrieved papers from CAR's corpus of up to $20$ papers for the keyword, The AI Scientist-v2 runs its own Semantic Scholar search, and The AI Scientist's ideation stage uses no literature. Every method runs on GPT-5.4 and GLM-5.2, and two numbers given without labels are for these backbones in this order. To lower cost, CAR's graph exploration agent uses DeepSeek-V4-Pro, and the ideation agent may send it only one natural-language query per idea. We generate up to five ideas per keyword.

\textbf{FML-bench and BioXArena.} CAR's proposals are supplied to the executor of The AI Scientist-v2, adapted to BioXArena's time limits (Appendix \ref{app:expsettings_bioxarena}).
We compare CAR with The AI Scientist-v2 using this shared
executor and with the methods reported by each benchmark. CAR and The AI Scientist-v2 use GPT-5.4, as do the FML-bench agents, while the BioXArena baselines keep the benchmark's own model specifications. We generate three ideas per run, with three runs on FML-bench and one on BioXArena, following each benchmark's protocol.

\section{Results and Analysis}
\label{results}

\subsection{Originality and feasibility}
\label{results_originality}

Table \ref{tab:main} reports the LiveIdeaBench dimensions and MN. We read originality and feasibility as the primary axes, since they trade off across methods and feasibility is what \S\ref{results_effectiveness} tests by execution.

\begin{table}[t]\centering\small\renewcommand{\arraystretch}{0.92}\setlength{\tabcolsep}{4pt}
\caption{Research ideation evaluation results on LiveIdeaBench (mean $\pm$ SE over 40 keywords). O = originality, F = feasibility, C = clarity, Flu. = fluency, MN = methodological novelty. $\uparrow$ indicates higher is better. \textbf{Bold} = best per backbone, \underline{underline} = second best. Flexibility: Appendix~\ref{app:liveideabench_full}. }
\label{tab:main}
\begin{tabular}{@{}lccccc@{}}
\toprule
Method & O $\uparrow$ & F $\uparrow$ & C $\uparrow$ & Flu. $\uparrow$ & MN $\uparrow$ \\
\midrule
\multicolumn{6}{c}{\textbf{GPT-5.4}} \\
\midrule
Base & $6.38\pm0.11$ & $4.38\pm0.10$ & $7.52\pm0.07$ & $6.15\pm0.29$ & $5.36\pm0.14$ \\
RIA & $\textbf{7.16}\pm0.07$ & $4.37\pm0.07$ & $7.06\pm0.07$ & $\underline{7.97}\pm0.16$ & $\textbf{6.88}\pm0.13$ \\
AI Scientist & $4.69\pm0.11$ & $\textbf{7.90}\pm0.09$ & $\textbf{7.80}\pm0.07$ & $6.19\pm0.20$ & $5.81\pm0.25$ \\
AI Scientist-v2 & $6.13\pm0.10$ & $\underline{6.02}\pm0.14$ & $\underline{7.72}\pm0.07$ & $7.11\pm0.17$ & $6.23\pm0.13$ \\
\textbf{CAR} (ours) & $\underline{6.75}\pm0.09$ & $4.97\pm0.11$ & $6.80\pm0.13$ & $\textbf{8.28}\pm0.15$ & $\underline{6.63}\pm0.13$ \\
\midrule
\multicolumn{6}{c}{\textbf{GLM-5.2}} \\
\midrule
Base & $5.60\pm0.08$ & $3.76\pm0.11$ & $6.99\pm0.10$ & $7.94\pm0.19$ & $4.87\pm0.19$ \\
RIA & $\underline{6.45}\pm0.10$ & $3.40\pm0.10$ & $6.42\pm0.10$ & $\underline{8.67}\pm0.10$ & $\underline{6.58}\pm0.14$ \\
AI Scientist & $5.66\pm0.09$ & $\textbf{6.29}\pm0.13$ & $\underline{7.53}\pm0.07$ & $6.88\pm0.18$ & $6.16\pm0.16$ \\
AI Scientist-v2 & $6.29\pm0.10$ & $\underline{5.39}\pm0.17$ & $\textbf{7.62}\pm0.08$ & $7.54\pm0.25$ & $6.44\pm0.20$ \\
\textbf{CAR} (ours) & $\textbf{6.89}\pm0.08$ & $4.59\pm0.09$ & $6.90\pm0.11$ & $\textbf{8.80}\pm0.14$ & $\textbf{6.74}\pm0.09$ \\
\bottomrule
\end{tabular}
\end{table}

\textbf{No method leads on both axes.} On GPT-5.4, The AI Scientist is the most feasible and least original method, and RIA, whose prompt solicits unconventional ideas, is the most original and least feasible. On GLM-5.2, The AI Scientist remains the most feasible and RIA the least feasible, while CAR is the most original (Appendix \ref{app:axes_liveideabench} plots the two axes). In the expert-rated proposals of \citet{si2025can}, higher originality is also associated with lower feasibility (Appendix \ref{app:axes}).
Research proposal types also differ: among $119$ ideas with complete
annotations, The AI Scientist mostly proposes studies and
resources, while RIA and Base propose methods almost
exclusively (Appendix \ref{app:ideatype}).

\textbf{CAR sits at the original end with more feasibility than RIA.} CAR is $0.41$ below RIA in originality on GPT-5.4 and $0.44$ above it on GLM-5.2, and $0.60$ and $1.19$ more feasible. Against The AI Scientist-v2, it gains about $0.6$ originality for about a point of feasibility on both backbones. CAR is non-dominated on both backbones and has the highest fluency of all methods: its five ideas for a keyword are the most distinct from one another. Its clarity is the lowest on GPT-5.4.

\textbf{Methodological novelty.} CAR's MN is $6.63$ on GPT-5.4 and $6.74$ on GLM-5.2. It is significantly above Base ($5.36$, $4.87$) and The AI Scientist ($5.81$, $6.16$) on both backbones (Holm-corrected $p<.01$ over the four comparisons within each backbone). It is above The AI Scientist-v2 ($6.23$, $6.44$) on both, but not significantly ($p=.058$, $.56$). It is below RIA on GPT-5.4 ($6.88$) and above it on GLM-5.2 ($6.58$), neither significantly ($p\geq.21$), so we do not claim that it leads on novelty. The contrasts with Base and The AI Scientist are positive in all four scientific domains on both backbones, with no significant heterogeneity (Appendix \ref{app:domains}, paired SEs in Appendix \ref{app:liveideabench_full}).

\subsection{Effectiveness}
\label{results_effectiveness}

\begin{table*}[!t]
\centering
\caption{Normalized test improvement over the baseline on the eight FML-bench-Lite tasks (per-task results are mean$\pm$std over three runs,
and mean results are the mean$\pm$std of the per-task means across tasks). Pairwise win rate follows Equation~\ref{eq:winrate}, with scores rounded to three decimals first and ties counted as a win for neither method. TAS = The AI Scientist, TAS-v2 = The AI Scientist-v2. \textbf{Bold} = best, \underline{underline} = second best.}
\label{table_fmlbenchlite_results}
\resizebox{\textwidth}{!}{%
\begin{tabular}{lccccccc}
\toprule
Task & TAS & TAS-v2 & AIDE & AIRA & autoresearch & OpenEvolve & \textbf{CAR} (ours) \\
\midrule
PyCIL & $0.030\pm.025$ & $\underline{0.064}\pm.004$ & $0.056\pm.005$ & $0.020\pm.007$ & $0.026\pm.007$ & $0.037\pm.021$ & $\mathbf{0.082}\pm.004$ \\
USB & $0.040\pm.011$ & $0.039\pm.008$ & $0.029\pm.008$ & $0.019\pm.005$ & $0.048\pm.029$ & $\underline{0.067}\pm.012$ & $\mathbf{0.074}\pm.020$ \\
DomainBed-CM & $0.071\pm.041$ & $\mathbf{0.245}\pm.182$ & $0.231\pm.198$ & $0.165\pm.211$ & $0.067\pm.053$ & $\underline{0.242}\pm.200$ & $0.178\pm.097$ \\
DomainBed-OH & $0.006\pm.004$ & $\mathbf{0.014}\pm.003$ & $0.011\pm.005$ & $0.012\pm.002$ & $0.005\pm.004$ & $\underline{0.013}\pm.005$ & $\mathbf{0.014}\pm.001$ \\
OpenOOD & $0.023\pm.033$ & $\underline{0.039}\pm.017$ & $0.012\pm.020$ & $0.021\pm.002$ & $\mathbf{0.057}\pm.013$ & $0.006\pm.011$ & $0.024\pm.020$ \\
Opacus & $0.000\pm.000$ & $0.000\pm.000$ & $0.006\pm.010$ & $0.000\pm.000$ & $\mathbf{0.054}\pm.017$ & $0.003\pm.003$ & $\underline{0.011}\pm.007$ \\
PrivacyMeter & $0.008\pm.014$ & $0.482\pm.068$ & $0.530\pm.018$ & $0.011\pm.020$ & $\mathbf{0.575}\pm.025$ & $0.032\pm.056$ & $\underline{0.542}\pm.021$ \\
ART & $0.429\pm.026$ & $0.395\pm.026$ & $0.447\pm.035$ & $0.303\pm.127$ & $\underline{0.448}\pm.032$ & $\mathbf{0.459}\pm.011$ & $0.411\pm.017$ \\
\midrule
Mean$\pm$std & $0.076\pm.144$ & $0.160\pm.190$ & $\underline{0.165}\pm.214$ & $0.069\pm.109$ & $0.160\pm.220$ & $0.107\pm.162$ & $\mathbf{0.167}\pm.202$ \\
Pairwise win rate & 27.1\% & 56.2\% & 50.0\% & 16.7\% & \underline{58.3\%} & \underline{58.3\%} & \textbf{75.0\%} \\
\bottomrule
\end{tabular}%
}
\end{table*}

\begin{table*}[!t]
\centering
\caption{Normalized task performance on the ten tasks of BioXArena's human-comparison pilot
(one run per method). The mean results are the mean$\pm$std across the nine valid tasks. cell-type-from-expression has a label leak that makes it trivially solvable and is shown for completeness but excluded from the mean and the pairwise win rate. Pairwise win rate follows Equation~\ref{eq:winrate}, with scores rounded to three decimals first and ties counted as a win for neither entry. TAS-v2 = The AI Scientist-v2. \textbf{Bold} = best, \underline{underline} = second best.}
\label{table_bioxarena_results}
\resizebox{\textwidth}{!}{%
\begin{tabular}{lccccccc}
\toprule
Task & STELLA & MLEvolve & Gemini-3.1-Pro & GPT-5.4 & TAS-v2 & \textbf{CAR} (ours) & Human \\
\midrule
cyp-inhibition-multi-label & 0.891 & 0.862 & 0.898 & \underline{0.919} & 0.877 & 0.897 & \textbf{0.929} \\
gpcr-binding-multi-class & 0.937 & 0.937 & 0.933 & 0.894 & 0.943 & \textbf{0.947} & \underline{0.945} \\
go-function-multi-label & 0.496 & 0.508 & \textbf{0.513} & 0.506 & \underline{0.510} & 0.507 & 0.499 \\
synthetic-lethality-prediction & 0.967 & \underline{0.971} & 0.966 & 0.958 & 0.970 & \underline{0.971} & \textbf{0.973} \\
cancer-drug-sensitivity & 0.959 & 0.972 & 0.956 & \underline{0.974} & 0.972 & \textbf{0.992} & 0.972 \\
breast-cancer-subtype & 0.684 & 0.601 & 0.704 & \textbf{0.732} & 0.719 & 0.624 & \underline{0.723} \\
pan-cancer-survival-prediction & 0.777 & \textbf{0.790} & 0.777 & 0.703 & 0.662 & 0.775 & \underline{0.785} \\
variant-effect-pathogenicity & 0.458 & 0.592 & 0.472 & 0.579 & \textbf{0.630} & \underline{0.607} & 0.582 \\
cell-type-from-expression & 0.918 & \underline{0.996} & \textbf{1.000} & \textbf{1.000} & \textbf{1.000} & \textbf{1.000} & \textbf{1.000} \\
gene-expression-classification & 0.960 & 0.945 & 0.961 & 0.950 & 0.968 & \underline{0.972} & \textbf{0.981} \\
\midrule
Mean$\pm$std & $0.792\pm.203$ & $0.798\pm.184$ & $0.798\pm.195$ & $0.802\pm.177$ & $0.806\pm.177$ & $\underline{0.810}\pm.187$ & $\mathbf{0.821}\pm.184$ \\
Pairwise win rate & 25.9\% & 40.7\% & 40.7\% & 40.7\% & 53.7\% & \underline{64.8\%} & \textbf{72.2\%} \\
\bottomrule
\end{tabular}%
}
\end{table*}

\textbf{FML-bench-Lite.} CAR attains the highest pairwise win rate ($75.0\%$, with next best method at $58.3\%$) and the highest mean normalized test improvement ($0.167$, with AIDE at $0.165$ and The AI Scientist-v2 and autoresearch at $0.160$). The top means are close and dominated by PrivacyMeter and ART, so pairwise win rate is the more informative metric. With the same executor, CAR attains a higher win rate than The AI Scientist-v2 ($75.0\%$ vs.\ $56.2\%$), though outcomes also depend on implementation and execution variability. CAR underperforms several agents on DomainBed-CM, OpenOOD and ART. On OpenOOD, the proposed training-time mechanism was not identified in the saved implementation
(Appendix \ref{app:cases}).

\textbf{BioXArena.} Without a task template, CAR obtains the highest win rate ($64.8\%$) and mean score ($0.810$) among AI methods, ahead of The AI Scientist-v2 with the same executor ($53.7\%$, $0.806$) and of the benchmark's AI baselines ($25.9$--$40.7\%$, $0.792$--$0.802$), while the human reference remains ahead ($72.2\%$, $0.821$).\footnote{Results use the original BioXArena release. The organizers' corrected release of 7 October 2026 rebuilt six of these tasks, and we will update these results in a later version.} With one run per task, mean differences of a few thousandths should be read with the win rate and tie counts (Appendix \ref{app:ties}).

On LiveIdeaBench, the judges rate CAR's ideas as less feasible than The AI Scientist-v2's (Table~\ref{tab:main}). On FML-bench-Lite and BioXArena, where the two share an executor, CAR's ideas win more pairwise comparisons. The executed ideas are generated for the two benchmarks' tasks and are not the ones the LiveIdeaBench judges rated, so this compares different sets of ideas rather than testing whether a low feasibility score predicts poor execution for the same idea.

\subsection{Ablation study}
\label{results_ablation}

\begin{table}[!t]\centering\small\setlength{\tabcolsep}{3pt}
\caption{Ablations on LiveIdeaBench ($40$ keywords). All conditions share CAR's corpus, entities and harness. $\Delta$MN: CAR's MN minus the row's, computed per keyword and averaged ($\pm$ SE over keywords). It equals the difference of the two MN columns apart from rounding. Clarity, fluency, standard errors for O, F and MN, and Holm-corrected $p$ values are in Appendix~\ref{app:liveideabench_full}.}
\label{tab:ablation}
\begin{tabular}{@{}lcccc@{\hspace{10pt}}cccc@{}}
\toprule
 & \multicolumn{4}{c}{GPT-5.4} & \multicolumn{4}{c}{GLM-5.2} \\ \cmidrule(r){2-5}\cmidrule(l){6-9}
Condition & O$\uparrow$ & F$\uparrow$ & MN$\uparrow$ & $\Delta$MN$\uparrow$ & O$\uparrow$ & F$\uparrow$ & MN$\uparrow$ & $\Delta$MN$\uparrow$ \\
\midrule
\textbf{CAR} (full) & 6.75 & 4.97 & 6.63 & -- & 6.89 & 4.59 & 6.74 & -- \\
\midrule
Floor: one step, no creativity prompt & 6.54 & 5.10 & 6.23 & $+0.40\pm0.15$ & 6.81 & 4.77 & 6.38 & $+0.36\pm0.12$ \\
Direct: one step, processes in prompt & 6.57 & 5.02 & 6.33 & $+0.31\pm0.14$ & 6.83 & 4.68 & 6.53 & $+0.20\pm0.12$ \\
Freeform: iterative, no operators & 6.69 & 5.08 & 6.66 & $-0.02\pm0.12$ & 6.90 & 4.67 & 6.36 & $+0.37\pm0.12$ \\
\midrule
KG relation-label shuffle & 6.67 & 5.06 & 6.67 & $-0.04\pm0.12$ & 6.73 & 4.68 & 6.55 & $+0.19\pm0.13$ \\
No operator validation & 6.56 & 5.05 & 6.68 & $-0.05\pm0.11$ & 6.80 & 4.81 & 6.64 & $+0.10\pm0.10$ \\
\midrule
No \textit{bisociate} & 6.55 & 5.23 & 6.44 & $+0.20\pm0.12$ & 6.72 & 4.72 & 6.63 & $+0.11\pm0.11$ \\
No \textit{analogize} & 6.64 & 4.93 & 6.69 & $-0.06\pm0.13$ & 6.77 & 4.57 & 6.71 & $+0.02\pm0.12$ \\
No \textit{explore} & 6.63 & 5.02 & 6.61 & $+0.03\pm0.13$ & 6.70 & 4.61 & 6.39 & $+0.34\pm0.14$ \\
No \textit{elaborate} & 6.57 & 4.98 & 6.65 & $-0.01\pm0.09$ & 6.84 & 4.46 & 6.43 & $+0.30\pm0.10$ \\
\bottomrule
\end{tabular}
\end{table}

Nine ablated conditions address three questions
(Table \ref{tab:ablation}):
\textbf{(RQ1)} do operators improve ideas beyond prompting
and free-form iteration,
\textbf{(RQ2)} do relation labels or validation affect quality,
and \textbf{(RQ3)} does any single operator matter?
No MN contrast survives Holm correction over the nine
conditions within each backbone (smallest $p=.076$).
We therefore discuss directional patterns rather than
established MN gains.

\textbf{Operators versus prompting (RQ1).}
\emph{Floor} proposes in one step without naming the creativity
processes, \emph{Direct} names them, and \emph{Freeform}
retains iterative reasoning and graph tools without structured
operators.
CAR's MN exceeds Floor by $0.40\pm0.15$ and $0.36\pm0.12$
and Direct by $0.31\pm0.14$ and $0.20\pm0.12$.
Both backbones thus favor CAR over one-step proposal in
their point estimates.
Against Freeform, the differences are $-0.02\pm0.12$ and
$+0.37\pm0.12$, with no significant critic contrast
(Holm $p\geq.30$).
These results do not establish an advantage from operators
beyond free-form iteration.

\textbf{Relation labels and validation (RQ2).}
Shuffling edge labels while retaining entity content and
topology gives CAR-minus-shuffle MN differences of
$-0.04\pm0.12$ and $+0.19\pm0.13$, with no significant
critic contrast.
Removing validation gives MN differences of
$-0.05\pm0.11$ and $+0.10\pm0.10$.
Against the no-validation condition, CAR's originality advantage
is significant only on GPT-5.4 ($+0.19$).
Validation rejects $218/1{,}656$ attempted operations.
Of these, $176$ are immediately followed by an accepted call
of the same operator, and $213$ have that operator as their
next successful call (Appendix \ref{app:traces}).
These checks establish structural validity, not scientific
correctness.

\textbf{Individual operators (RQ3).}
On GPT-5.4, no removal changes MN by more than $0.20$.
On GLM-5.2, CAR exceeds the conditions without
\textit{explore} and \textit{elaborate} by $0.34\pm0.14$
and $0.30\pm0.10$, respectively, and the other removals by
at most $0.11$.
No individual operator has an established MN benefit
across both backbones.

\textbf{Summary.}
Across both backbones, MN point estimates favor CAR over
one-step proposal, while the operators make its development
inspectable. In $1{,}600$ audited ideas, all but one of
$16{,}820$ declared entity indices appear in the preceding
logs (Appendix \ref{app:traces}).

\subsection{Case study}
\label{results_case}

\begin{figure}[!t]
\centering
\includegraphics[width=\linewidth]{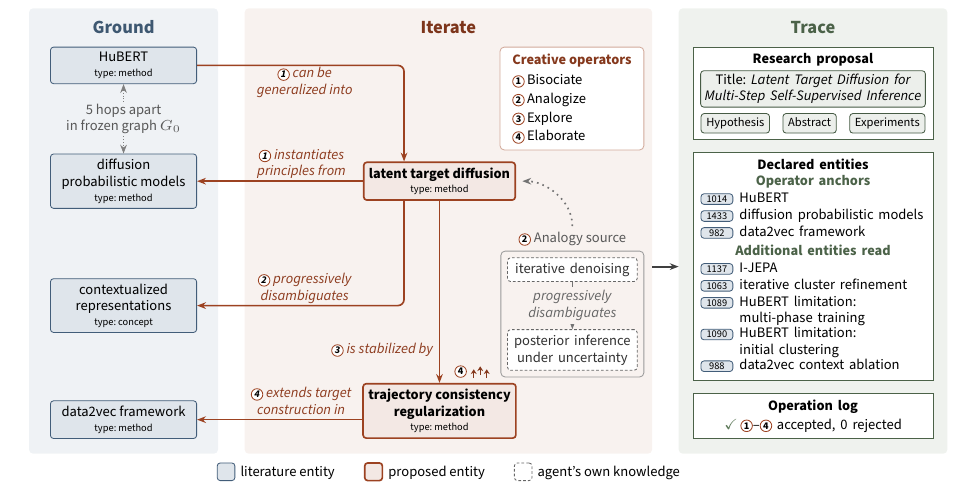}
\caption{A CAR trace for \emph{unsupervised learning} generated using GPT-5.4.
The proposal declares three of four operator anchors and
five additional literature entities read through graph tools. Entity names are shortened for space.}
\label{fig:case}
\end{figure}

Figure~\ref{fig:case} traces an
\emph{unsupervised learning} proposal.
\textit{Bisociate} bridges \emph{HuBERT} and
\emph{diffusion probabilistic models} through
\emph{latent target diffusion}.
\textit{Analogize} transfers ``progressively disambiguates''
from iterative denoising and posterior inference to latent
target diffusion and contextualized representations.
\textit{Explore} adds \emph{trajectory consistency regularization},
which \textit{Elaborate} refines with a link to
\emph{data2vec}. Appendix~\ref{app:cases} presents an execution example and a failure case.

\section{Conclusion}
\label{sec:conclusion}

CAR combines literature grounding, iterative development and
explicit creative operators for research ideation. It exceeds
single-shot prompting and The AI Scientist in methodological
novelty and achieves the highest win rate among the evaluated
AI methods on both execution benchmarks. MN point estimates favor CAR over one-step proposal on both backbones, although no MN ablation contrast survives correction for multiple comparisons.
The operators make idea development inspectable, but their
quality benefit beyond free-form iteration is not established.

\textbf{Limitations.}
Traces establish procedural provenance, not causal necessity.
Judge agreement is assessed on proposals used in MN evaluator
development, without human quality ratings of our ideas.
Retrieval, extraction errors and limited execution tasks
constrain the findings (Appendix \ref{app:limitations}).

\clearpage %
\subsection*{AI use statement}

In this work, we used generative AI tools to design or provide feedback on research methodology or experiments, implement methods, clean and reformat datasets, support qualitative and thematic data analysis, and interpret results. We have not used generative AI tools to generate synthetic data sets, help develop theoretical models or conceptual frameworks or propose or refine hypotheses, and the remaining required disclosure tasks, namely to formulate mathematical claims, provide critical ingredients for proving mathematical claims, assist in the writing of proofs, and assist with translation, are not applicable to this work. The statement above describes how we used AI to carry out
this study and write the paper. Separately, generating research
proposals is the experimental task performed by CAR and
the comparison methods. Additionally, we used generative AI tools to formulate questions for surveys or interviews, create or modify scientific figures or images, suggest experimental parameters, create or edit software code, create artifacts, draft parts of a research paper, summarize or analyze existing literature, discover research topics or identify gaps, brainstorm, source or search for information, edit a research paper to improve readability, identify relevant literature, format references, suggest a structure for a research paper, and propose a title or keywords for a research paper.
We have reviewed all AI-assisted work. We take responsibility for checking AI-assisted text, code and analyses against the underlying sources and experimental records. AI-assisted code was reviewed and tested by the authors. AI-drafted text and literature summaries were checked against the cited papers by the authors, and the research-proposal-type labels in Appendix \ref{app:ideatype} are human labels. We take responsibility for the final content of this work, including text, claims and artifacts produced with the aid of generative AI.

\subsection*{Ethics statement}
This work automates part of research ideation. Generated ideas are untested proposals and may be wrong, infeasible or unsafe if pursued. CAR records which literature entities each idea builds on, which supports attribution to prior work, but it does not replace a literature review by the researcher. A small number of generated biomedical ideas were refused by one judge's provider content filter and are excluded and counted (Appendix \ref{app:coverage}).
The literature corpora are built from papers obtained as arXiv sources, as open-access PDFs, or under the authors' institutional subscriptions. Full texts are used only for entity and relation extraction and are not redistributed. We release paper identifiers, the extracted knowledge graphs, prompts and derived data, not the papers themselves.
The human ratings we use for instrument validation are the publicly released data of \citet{si2025can}. The idea-type labels (Appendix \ref{app:ideatype}) were provided by three members of the authors' group who labeled anonymized abstracts.

\subsection*{Reproducibility statement}

The main prompts are given in Appendix \ref{app:prompts}, baseline fidelity and output normalization in Appendix \ref{app:fidelity}, and models, providers, cut-off dates, decoding settings and keyword lists in Appendix \ref{app:expsettings}. We will release code, the per-keyword paper identifiers and knowledge graphs, all generated ideas with their operator traces, all judge outputs, and the scripts that regenerate every table from these files. Hosted LLMs are not deterministic, so we pin providers for open-weight models and report the variation across three independent end-to-end draws of the methodological novelty instrument (Appendix \ref{app:validation}). The number of ideas behind each MN result, and the ideas excluded for missing retrieval evidence or judge refusals, are given in Appendix \ref{app:coverage}.

\bibliographystyle{iclr2027_conference}
\bibliography{iclr2027_conference}

\appendix

\section{Knowledge graph repair and entity resolution}
\label{app:repair}

Per-paper extraction can leave scientific entities without relations
to other non-paper entities and can represent the same entity under
different names. We apply relation repair within each paper and
entity resolution across the assembled graph.

\textbf{Relation repair (within a paper).}
We measure \emph{relation coverage} as the fraction of non-paper
nodes incident to at least one edge whose endpoints are both
non-paper nodes. Edges incident to paper nodes record provenance
and are excluded from this measure. When coverage falls below
$80\%$, an LLM proposes additional relations using the extracted
entity summaries and existing relations, with up to 20 currently
unconnected entities highlighted in the prompt
(Appendix~\ref{app:prompts_knowledge_repair}).
A candidate relation is accepted only if both endpoints resolve
to entities already extracted from that paper, its relation label
and description are present, and its subject--relation--object
triple is not already present. This step adds relations without
adding entities. It does not guarantee that coverage reaches
the threshold.

\textbf{Entity resolution (across papers).}
Entities denoting the same concept can appear as distinct nodes.
Once all papers are inserted, we lowercase their names, trim
surrounding whitespace and remove punctuation. Nodes with matching
normalized names and the same entity type are merged. This resolves
surface-form variants but can leave synonyms with different names
as separate nodes.

Neither step introduces new entities: relation repair changes
connectivity, while entity resolution consolidates existing
nodes. These operations do not independently verify the
scientific correctness of the extracted entities or relations.

\section{Baseline implementations}
\label{app:fidelity}

We retain each baseline's native output format and prompting
structure, with the adaptations described below. Ideas are
normalized only after generation by separate LLM conversions
instructed to preserve their content. We do not change generation
prompts to require a shared output schema. Method-specific scaffold
fields, such as self-assigned ratings, are stripped before
conversion. The normalization procedure and generation budgets
are described in Appendix~\ref{app:expsettings}, and the prompts
are given in Appendix~\ref{app:prompts}.

\subsection{Base}
Each idea is generated by a single call with the original
LiveIdeaBench idea prompt and the keyword verbatim. There is
no retrieval and no memory of earlier ideas.

\subsection{Research Ideation Agent}
We use the idea generation prompts of \citet{si2025can} in their
native \emph{batched} form. Seed ideas for a keyword are produced
in two joint calls, one with retrieved papers and one without.
For five ideas, the allocation is three retrieved-paper ideas
and two plain-generation ideas, or the reverse, alternating by
the keyword's position in the sorted keyword list. Each seed
idea is then expanded into a full proposal using the method's
experiment-plan prompt. This expansion is part of generation,
before the shared evaluation conversions.

We omit the upstream over-generation, deduplication and ranking
stages and instead request five ideas per keyword. This matches
the target number of evaluated ideas, rather than the upstream
selection procedure or generation cost. The prompts'
NLP-specific wording is adapted to general scientific keywords,
while their demonstration examples are retained.

\subsection{The AI Scientist}
Ideas are generated sequentially with the method's prompting
structure and native fields: Name, Title, Experiment,
Interestingness, Feasibility and Novelty. Each new idea is
conditioned on the previous ideas in full and can be revised
within the configured iteration budget. The native outputs
are then converted for evaluation.

\subsection{The AI Scientist-v2}
We retain the method's ideation loop and native proposal format,
which CAR also uses. The iteration budget is adapted for our
comparison, as described in Appendix~\ref{app:expsettings}.
On GLM-5.2, it finalized $153$ of $200$ requested ideas.
Of the remaining $47$ attempts, $44$ used all three rounds
for literature searches without finalizing a proposal,
and three reached the round limit after response-parsing
failures. Missing ideas are not imputed.

\section{Experimental settings}
\label{app:expsettings}

We convert ideas to the formats expected by each evaluation
instrument. For the LiveIdeaBench critics, a separate LLM step
rewrites each idea with a target of 100 words
(Appendix~\ref{app:prompts_liveideabenchoriginal}).
Some converted texts exceed this target.
For the methodological novelty (MN) judge, a fixed converter
rewrites each idea as an abstract of approximately 250 words
(Appendix~\ref{app:prompts_overlapjudge}).

\textbf{Models and generation counts.}
We evaluate the LiveIdeaBench methods with GPT-5.4 and GLM-5.2.
Our CAR and The AI Scientist-v2 runs on FML-bench and BioXArena
use GPT-5.4. Other benchmark entries retain the models and
settings of the published comparison.
CAR's graph exploration agent uses DeepSeek-V4-Pro.
We request five ideas per keyword, method and backbone on
LiveIdeaBench. Missing outputs are reported in
Appendix~\ref{app:coverage}.
For CAR and The AI Scientist-v2 on the execution benchmarks,
we request three ideas per run, with three runs per task on
FML-bench and one on BioXArena.

\textbf{Converters.}
The 100-word conversion uses the generator's backbone.
The MN conversion uses GPT-5.4 for both backbones.
Newly staged MN conversions use temperature $0$, an output
limit of $900$ tokens, and seed $42$ plus the item's position
in the sorted manifest. Existing conversions are reused when
their source identity and converter provenance can be
established. Their provenance records distinguish saved
configurations from reconstructed historical settings.

\textbf{Providers.}
For MN evaluation, the campaign configuration pins the
DeepSeek judge to the first-party DeepSeek provider.
The GLM generation launch scripts specify Z.AI.
For CAR's ideation agent, these scripts specify
\texttt{tool\_choice=auto} for GLM-5.2 and
\texttt{tool\_choice=required} for GPT-5.4.

\begin{table}[t]
\centering
\small
\caption{Selected generation and evaluation settings for
LiveIdeaBench. Unspecified temperature uses the provider
default. Recorded baseline settings agree across both backbones.}
\label{tab:decoding_settings}
\begin{tabular}{lll}
\toprule
Stage & Temperature & Other settings \\
\midrule
RIA seed generation & $1.0$ & top-$p=1.0$, $30{,}000$ tokens \\
RIA proposal expansion & $0$ & $30{,}000$ tokens \\
The AI Scientist & $0.75$ & up to three total rounds \\
The AI Scientist-v2 & $0.7$ & up to three total rounds \\
100-word conversion & $0$ & target of 100 words \\
New MN conversions & $0$ & $900$ tokens \\
MN retrieval and judging & $0$ & up to 11 retrieval rounds \\
Originality, feasibility, clarity & Unspecified & fixed three-judge panel \\
Fluency & Unspecified & all available idea pairs \\
\bottomrule
\end{tabular}
\end{table}

\textbf{Corpus.}
We build one corpus of up to $20$ obtainable papers and its
graph per keyword or task, then reuse it across CAR runs.
For LiveIdeaBench, RIA's retrieval-augmented batch draws from
the same corpus. Base and The AI Scientist do not use this
corpus, while The AI Scientist-v2 performs its own literature
search. Of the $40$ LiveIdeaBench keywords, $38$ have $20$
papers. \emph{Cellular automata} has $12$ and
\emph{decision theory} has $17$, where fewer sources could
be obtained.

\textbf{Literature cut-offs.}
We set the publication-date cut-off for LiveIdeaBench corpus
retrieval and MN evaluation to 2026-06-16.
For our FML-bench and BioXArena literature retrieval, the
cut-offs are 2026-05-17 and 2026-05-15, respectively.
For MN validation on the data of \citet{si2025can}, the cut-off
is 2024-06-30.
These dates restrict retrieved evidence, not the knowledge
encoded in the models' parameters.

\textbf{MN retrieval.}
For each idea, we query Semantic Scholar for up to $11$ rounds.
Retrieval stops earlier if the candidate bank has reached
$120$ papers before the next round. Papers are deduplicated
by identifier and ranked by LLM relevance, with ties broken
by SPECTER2 similarity. The top $10$ papers form the evidence
for judging. Coverage and exclusions are reported in
Appendix~\ref{app:coverage}.

\subsection{LiveIdeaBench}
\label{app:expsettings_liveideabench}

The $40$ selected keywords are shown in
Table~\ref{liveideabench_keywords}.
We use a fixed three-judge panel of Claude-Sonnet-4.6,
Gemini-3.1-Pro and DeepSeek-V4-Pro instead of the benchmark's
dynamically selected panel. GPT-5.4 is excluded from the panel
to reduce the risk of self-preference when judging its outputs
\citep{wataoka2024self}.
We compute fluency by averaging over all pairs of available
ideas for a keyword and method, each judged by all three
critics. With five ideas, this gives $\binom{5}{2}=10$ pairs,
rather than a single randomly selected pair.
Critic and fluency calls leave temperature unspecified,
using the provider's default.

The AI Scientist and The AI Scientist-v2 allow up to three
rounds per idea, including the initial generation round.
They can therefore take at most two subsequent refinement
rounds and may stop earlier. CAR is instructed to complete at least
three accepted creative operations before submitting a
proposal. These settings do not match the methods in
refinement steps, model calls or computational cost.

\subsection{FML-bench}
\label{app:expsettings_fmlbench}

We use The AI Scientist-v2 executor as ported in the
FML-bench release, with $100$ steps split across baseline,
tuning, creative and ablation stages in proportions
$10/20/50/20\%$. Each execution has a time limit of
$2{,}400$\,s. CAR's three ideas are supplied through the
executor's external-ideas input. CAR and The AI Scientist-v2
use the same execution configuration.

For each run, we compute improvement over the task baseline
in the direction preferred by the metric, divide by the
benchmark's metric range, and clip negative improvements
to zero. For the eight FML-bench-Lite tasks, the range is
$1$ except for PrivacyMeter, where it is $0.5$ and lower
scores are better.

\subsection{BioXArena}
\label{app:expsettings_bioxarena}

The executor uses two stages under a nominal two-hour task
budget that includes idea generation. The execution search
receives the remaining budget after accounting for ideation.
Solutions are instructed to run within $10$ minutes.
To accommodate contention between three concurrent nodes,
the development harness permits up to $30$ minutes per
execution. Final-test execution retains the $10$-minute limit.

Stage one constructs a baseline and a validation split shared
by later nodes. Stage two creates three draft nodes in parallel
(\texttt{num\_parallel}$=3$), one per generated idea.
Two adaptations apply to both CAR and The AI Scientist-v2:
a richer data overview containing column names, value counts
and summary statistics, and a guard against using validation
labels for training.

In cell-type-from-expression, the released \texttt{train.csv}
and \texttt{test.csv} contain a \texttt{cluster} column whose
seven values are the cell-type names themselves.
The mapping has purity $1.0$ on all $3{,}252$ training rows,
and every test cluster appears in training.
A lookup reproduces CAR's submission on all $813$ test rows
and scores $1.000$. Two separate CAR executions with the same
seed produced byte-identical submissions.
STELLA and MLEvolve score $0.918$ and $0.996$, respectively,
while CAR, The AI Scientist-v2, the human entry and both
bare-LLM entries score $1.000$.
We retain the task in the per-task table but exclude it from
all aggregates because the leaked labels make it unsuitable
for comparing predictive performance.

We use the benchmark's normalized test scores.
For cancer-drug sensitivity, Spearman correlation $r$ is
mapped to $(r+1)/2$. The remaining pilot-task metrics
already lie in $[0,1]$.

\subsection{Rounding, ties and win rates}
\label{app:ties}
Following \citet{zou2026fml}, we write pairwise win rate as
\begin{equation}
\mathrm{WR}(A)
=
\frac{1}{(K-1)N}
\sum_{\substack{B\in\mathcal{A}\\B\neq A}}
\sum_{t=1}^{N}
\mathbf{1}\!\left[A\succ_t B\right],
\label{eq:winrate}
\end{equation}
where $\mathcal{A}$ is the comparison set,
$K=|\mathcal{A}|$ and $N$ is the number of tasks.
In our comparison, $A\succ_t B$ means that $A$ has a
strictly higher reported score than $B$ on task $t$
after rounding to three decimals.
Ties count as wins for neither entry. The comparison set $\mathcal{A}$ includes all reported entries,
including the human reference on BioXArena.

After rounding every agent's score to three decimals, $4$ of
the $168$ agent pairs on FML-bench-Lite and $6$ of the $189$
on BioXArena are tied (Table~\ref{tab:ties}).
One tie on each benchmark involves CAR.
The excluded cell-type row, where five entries score $1.000$,
would add $10$ tied pairs.

\begin{table}[h]\centering\small\setlength{\tabcolsep}{4pt}
\caption{Ties after rounding all entries' scores to three decimals.
A tie counts as a win for neither agent (Eq.~\ref{eq:winrate}).
Tied entries receive the average of the ranks they span.}
\label{tab:ties}
\begin{tabular}{llllc}
\toprule
Benchmark & Task & Tied entries & Value & Tied pairs (with CAR)\\
\midrule
FML-bench-Lite & DomainBed-OH & TAS-v2, CAR & 0.014 & 1 (1)\\
 & Opacus & TAS, TAS-v2, AIRA & 0.000 & 3 (0)\\
 & \emph{Total} & & & 4 of 168 (1)\\
\midrule
BioXArena (9 tasks) & gpcr-binding & STELLA, MLEvolve & 0.937 & 1 (0)\\
 & synthetic-lethality & MLEvolve, CAR & 0.971 & 1 (1)\\
 & cancer-drug-sensitivity & MLEvolve, TAS-v2, Human & 0.972 & 3 (0)\\
 & pan-cancer-survival & STELLA, Gemini-3.1-Pro & 0.777 & 1 (0)\\
 & \emph{Total} & & & 6 of 189 (1)\\
\bottomrule
\end{tabular}
\end{table}

\begin{table*}[t]
\centering
\small
\caption{Selected LiveIdeaBench keywords by domain.}
\label{liveideabench_keywords}
\begin{tabularx}{\textwidth}{@{}l>{\raggedright\arraybackslash}X@{}}
\hline
Scientific domain & Keywords \\
\hline
Computational sciences & reinforcement learning, stochastic processes, unsupervised learning, support vector machines, numerical analysis, information technology, sampling, mathematics, complex analysis, cellular automata \\
Life sciences & neuroscience, precision medicine, structural biology, antibiotic stewardship, geriatrics, natural selection, protein engineering, pandemic preparedness, surgical robotics, behavioral ecology \\
Physical sciences & computational chemistry, quantum optics, statistical physics, green chemistry, radioisotopes, seafloor spreading, particle physics, mtheory, raman spectroscopy, mineral identification \\
Social sciences & cognitive psychology, environmental economics, social networks, social psychology, developmental psychology, evolutionary psychology, educational psychology, decision theory, disaster management, research funding \\
\hline
\end{tabularx}%
\end{table*}

\section{Evaluation validation}
\label{app:validation}

\subsection{Originality and feasibility}
\label{app:validation_of}

We assess agreement with expert ratings using 98 proposals
from \citet{si2025can}'s Human Study Data: 49 AI-generated and
49 human-written ideas across seven topics, excluding the
reranked condition. There are 228 ratings in total.
Of the 98 ideas, 67 have two ratings, 30 have three, and one
has four.

\textbf{Human reference.}
We estimate agreement between reviewers by split-half
resampling. For each of 2,000 random splits, with seed 2026,
we shuffle each idea's ratings and assign $\lfloor k/2\rfloor$
of its $k$ ratings to the first half and the remainder to
the second. We average ratings within each half and compute
the Spearman correlation between the two vectors of
idea-level scores. The mean split-half correlation for
novelty is $0.262$.

Applying the Spearman--Brown adjustment,
$\rho_{\mathrm{adj}}=2\rho/(1+\rho)$, gives a reference
value of $0.415$, with a basic bootstrap 95\% interval of
$[0.113,0.605]$ over ideas. This is an extrapolated reliability
reference rather than observed agreement between two
independent full panels. Its interpretation is approximate
because the halves can be unequal, the number of ratings
varies across ideas, and the adjustment is applied to
Spearman correlations.

\textbf{Critic agreement.}
We repeat conversion and scoring three times for each
converter backbone. Within each draw, the panel score is
the mean of the three critics' scores for an idea.
Tables~\ref{tab:validation_originality}
and~\ref{tab:validation_feasibility} report agreement with
the corresponding mean expert ratings.
The originality panel reaches mean Spearman correlations
of $0.362$ with GPT-5.4 conversion and $0.383$ with GLM-5.2
conversion. Both point estimates are below the adjusted
human reference. These comparisons do not establish
equivalence to human reviewers.

The feasibility panel reaches $0.379$ and $0.328$.
The reviewers' raw split-half feasibility correlation is
$0.123$. Its adjusted reference is $0.219$, with a 95\%
interval of $[-0.172,0.472]$. We therefore report feasibility
correlations as convergent evidence without a difference
test against that reference.

\begin{table}[t]
\centering
\small
\begin{tabular}{lcccc}
\toprule
& \multicolumn{2}{c}{GPT-5.4 converter}
& \multicolumn{2}{c}{GLM-5.2 converter} \\
\cmidrule(lr){2-3}\cmidrule(lr){4-5}
Judge & $\rho$ & $p_{\max}$ & $\rho$ & $p_{\max}$ \\
\midrule
Claude Sonnet 4.6
& $0.375\pm0.032$ & .0003
& $0.382\pm0.046$ & .0007 \\
Gemini 3.1 Pro
& $0.303\pm0.057$ & .007
& $0.340\pm0.088$ & .017 \\
DeepSeek-V4-Pro
& $0.219\pm0.028$ & .053
& $0.222\pm0.106$ & .305 \\
\midrule
Panel
& $0.362\pm0.017$ & .0005
& $0.383\pm0.023$ & .0003 \\
\bottomrule
\end{tabular}
\caption{Originality critics against expert novelty ratings
on 98 ideas. Values are mean idea-level Spearman correlations
across three conversion-and-scoring draws.
$\pm$ denotes the standard deviation (SD) across draws, not a sampling SE.
$p_{\max}$ is the largest correlation-test $p$-value
across the three draws.}
\label{tab:validation_originality}
\end{table}

\begin{table}[t]
\centering
\small
\begin{tabular}{lcccc}
\toprule
& \multicolumn{2}{c}{GPT-5.4 converter}
& \multicolumn{2}{c}{GLM-5.2 converter} \\
\cmidrule(lr){2-3}\cmidrule(lr){4-5}
Judge & $\rho$ & $p_{\max}$ & $\rho$ & $p_{\max}$ \\
\midrule
Claude Sonnet 4.6
& $0.340\pm0.035$ & .003
& $0.219\pm0.048$ & .099 \\
Gemini 3.1 Pro
& $0.325\pm0.035$ & .004
& $0.274\pm0.035$ & .012 \\
DeepSeek-V4-Pro
& $0.299\pm0.095$ & .053
& $0.347\pm0.037$ & .002 \\
\midrule
Panel
& $0.379\pm0.059$ & .002
& $0.328\pm0.015$ & .002 \\
\bottomrule
\end{tabular}
\caption{Feasibility critics against expert feasibility ratings
on the same 98 ideas. Values are mean idea-level Spearman
correlations across three conversion-and-scoring draws.
$\pm$ denotes the SD across draws.
$p_{\max}$ is the largest correlation-test $p$-value
across the three draws.}
\label{tab:validation_feasibility}
\end{table}

\subsection{Methodological novelty judge}
\label{app:validation_mn}

\textbf{Agreement with expert ratings.}
Across three draws of conversion, retrieval, ranking and
scoring, the MN panel reaches mean idea-level Spearman
$\rho=0.391\pm0.017$ on the same 98 proposals
(Table~\ref{tab:validation_mn}).

Averaging the three draws' scores for each idea before
computing the correlation gives $\rho=0.407$.
This is a different statistic from the mean of the three
draw-wise correlations. Its difference from the adjusted
human reference is $\Delta=-0.008$.
A bootstrap over ideas, with the human reference re-estimated
in each resample, gives a basic 95\% interval of
$[-0.30,+0.25]$. The calculation uses 2,000 bootstrap
resamples with seed 2026 and 50 reviewer splits within
each resample. The wide interval does not establish
equivalence to human reviewers.

These results measure agreement on the selected NLP proposals.
This sample also informed the choice of retrieval and ranking
configuration, so the reported agreement is a development-set
assessment rather than independent held-out validation.
Repeated pipeline draws assess variability on the same proposals,
not generalization to new ideas. They do not establish agreement
with experts across the 40 multidisciplinary LiveIdeaBench keywords.

\begin{table}[t]
\centering
\small
\begin{tabular}{lcc}
\toprule
Judge & $\rho$ & $p_{\max}$ \\
\midrule
Claude Sonnet 4.6 & $0.390\pm0.052$ & .0008 \\
Gemini 3.1 Pro & $0.339\pm0.035$ & .0023 \\
DeepSeek-V4-Pro & $0.290\pm0.073$ & .042 \\
\midrule
Panel & $0.391\pm0.017$ & .0002 \\
\bottomrule
\end{tabular}
\caption{Methodological novelty judges against expert novelty
ratings on 98 ideas. Values are mean idea-level Spearman
correlations across three draws of conversion, retrieval,
ranking and scoring. $\pm$ denotes the SD across draws.
$p_{\max}$ is the largest correlation-test $p$-value
across the three draws. The separate draw-averaged-score
comparison is reported in the text.}
\label{tab:validation_mn}
\end{table}

\textbf{Overlap with the generation corpus.}
The MN judge retrieves evidence from Semantic Scholar
without receiving CAR's graph or corpus list.
We measure how often its retrieved papers overlap with
each keyword's frozen generation corpus.
Across CAR, Freeform and Direct on both backbones,
approximately $9$--$13\%$ of the retrieved paper slots
shown to the judges contain corpus papers.
At least one corpus paper appears for $59$--$68\%$ of
ideas, and the full retrieved pool covers $9$--$12\%$
of the generation corpus.

For the four comparison baselines, the corresponding
ranges are $0.4$--$1.9\%$ of judged paper slots,
$4$--$12\%$ of ideas with any overlap, and $1$--$3\%$
of the generation corpus recovered.
RIA's retrieval-augmented batch has access to this corpus.
These descriptive overlap rates do not establish the
direction or magnitude of any effect on MN scores.

\textbf{Agreement between instruments.}
We compare MN and the originality, feasibility and clarity
critics on 2,708 ideas with a complete MN panel and available
critic scores. This analysis includes the five main methods,
Direct and Freeform on both backbones.
MN and originality correlate at Spearman $\rho=0.35$
across ideas, $0.44$ across keyword--method--backbone cells
($n=560$), and $0.79$ across method--backbone means ($n=14$).
The corresponding correlations with clarity are
$-0.08$, $-0.09$ and $-0.56$, and those with feasibility
are $-0.09$, $-0.06$ and $-0.24$.
MN is therefore positively associated with the originality
critic and only weakly associated with clarity and
feasibility at the idea level. These associations do not
isolate effects of writing style, method or scientific domain.

\section{Full results and evaluation coverage}
\label{app:full_results}

\subsection{MN coverage}
\label{app:coverage}
Table~\ref{tab:mn_coverage} reports main-method MN coverage.
Ideas are excluded, never imputed, when retrieval returns no
eligible paper after all $11$ rounds or when retrieval,
relevance scoring or judging fails to produce a valid complete
panel. Finding no eligible paper is a limitation of the retrieval
procedure, not proof that no related work exists.

Across the main methods, $1{,}925$ of $1{,}953$ generated ideas
have complete three-judge panels. Six ideas are excluded for
insufficient evidence and $22$ for technical failures, including
nine provider refusals. The AI Scientist-v2 produced $153$ of
$200$ requested ideas on GLM-5.2 (Appendix~\ref{app:fidelity}).
Its $47$ missing generations are outside these coverage counts.
Across the ten ablation conditions, including CAR, $3{,}928$ of
$4{,}000$ generated ideas have complete panels, with six
insufficient-evidence and $66$ technical exclusions.
CAR appears in both summaries.

Absolute MN columns use all ideas with a complete panel,
averaged within keyword and then across keywords. The paired
contrasts of Tables~\ref{tab:matched_mn_main}
and~\ref{tab:matched_mn_ablations} take differences of these
means for the same keywords, without matching individual idea
slots. Each contrast therefore equals the difference of the
two absolute columns up to rounding.

\begin{table}[t]
\centering\small
\caption{MN coverage of the main methods: ideas with a complete
three-judge panel over ideas generated. Evidence denotes no
eligible papers after $11$ retrieval rounds. Technical denotes
retrieval, relevance-scoring or judging failures, including
provider refusals. Excluded ideas are not imputed.}
\label{tab:mn_coverage}
\begin{tabular}{@{}llrrrr@{}}
\toprule
Backbone & Method & Scored/available & Evidence & Technical & Pending \\
\midrule
GPT-5.4 & Base & 199/200 & 0 & 1 & 0 \\
GPT-5.4 & Research Ideation Agent & 198/200 & 0 & 2 & 0 \\
GPT-5.4 & AI Scientist & 197/200 & 0 & 3 & 0 \\
GPT-5.4 & AI Scientist-v2 & 197/200 & 2 & 1 & 0 \\
GPT-5.4 & CAR (ours) & 197/200 & 1 & 2 & 0 \\
GLM-5.2 & Base & 195/200 & 0 & 5 & 0 \\
GLM-5.2 & Research Ideation Agent & 199/200 & 0 & 1 & 0 \\
GLM-5.2 & AI Scientist & 197/200 & 2 & 1 & 0 \\
GLM-5.2 & AI Scientist-v2 & 151/153 & 0 & 2 & 0 \\
GLM-5.2 & CAR (ours) & 195/200 & 1 & 4 & 0 \\
\bottomrule
\end{tabular}
\end{table}

\subsection{Full LiveIdeaBench tables}
\label{app:liveideabench_full}
Tables~\ref{tab:main} and~\ref{tab:ablation} in the main text
are reduced views of the tables below, which add standard
errors for every cell, clarity and fluency for the ablations,
and flexibility. Flexibility is the 30th percentile across
keywords of the within-keyword average of originality,
feasibility, clarity and fluency. Its standard error is estimated
from $20{,}000$ bootstrap resamples of keywords, with seed $42$.
The other metrics report standard errors across keyword means.
We report flexibility for completeness and interpret it
cautiously given the $40$-keyword sample.

\begin{table}[t]
\centering
\small
\setlength{\tabcolsep}{4pt}
\caption{August/September 2026 native-generation results. O: originality, F: feasibility, C: clarity, Flu.: fluency. O/F/C/Flu. report mean $\pm$ SE across 40 equally weighted keywords. Flex. is the 30th percentile of keyword mean O/F/C/Flu., with bootstrap SE (20,000 resamples). MN is $10-$mean methodological overlap across three judges, averaged within keyword then across available keywords. Errors are keyword SE. MN uses every idea with a complete panel. The paired contrasts of Table~\ref{tab:matched_mn_main} are differences of these keyword means. Missing generations, insufficient evidence and technical failures are not imputed. This snapshot includes separately preserved serial results and is not a fully completed MN evaluation. MN coverage: 1925/1953 available ideas, with 6 insufficient-evidence exclusions, 22 technical/provider failures (including 9 refusals), 0 pending. GLM AI Scientist-v2 has 47 missing native generations, excluded from its 153-idea denominator.}
\label{tab:main_liveideabench_available}
\begin{tabular}{@{}lcccccc@{}}
\toprule
Method & O $\uparrow$ & F $\uparrow$ & C $\uparrow$ & Flu. $\uparrow$ & Flex. $\uparrow$ & MN $\uparrow$ \\
\midrule
\multicolumn{7}{c}{\textbf{GPT-5.4}} \\
\midrule
Base & $6.38\pm0.11$ & $4.38\pm0.10$ & $7.52\pm0.07$ & $6.15\pm0.29$ & $5.80\pm0.14$ & $5.36\pm0.14$ \\
Research Ideation Agent & $7.16\pm0.07$ & $4.37\pm0.07$ & $7.06\pm0.07$ & $7.97\pm0.16$ & $6.45\pm0.06$ & $6.88\pm0.13$ \\
AI Scientist & $4.69\pm0.11$ & $7.90\pm0.09$ & $7.80\pm0.07$ & $6.19\pm0.20$ & $6.53\pm0.07$ & $5.81\pm0.25$ \\
AI Scientist-v2 & $6.13\pm0.10$ & $6.02\pm0.14$ & $7.72\pm0.07$ & $7.11\pm0.17$ & $6.59\pm0.07$ & $6.23\pm0.13$ \\
CAR (ours) & $6.75\pm0.09$ & $4.97\pm0.11$ & $6.80\pm0.13$ & $8.28\pm0.15$ & $6.49\pm0.07$ & $6.63\pm0.13$ \\
\midrule
\multicolumn{7}{c}{\textbf{GLM-5.2}} \\
\midrule
Base & $5.60\pm0.08$ & $3.76\pm0.11$ & $6.99\pm0.10$ & $7.94\pm0.19$ & $5.84\pm0.10$ & $4.87\pm0.19$ \\
Research Ideation Agent & $6.45\pm0.10$ & $3.40\pm0.10$ & $6.42\pm0.10$ & $8.67\pm0.10$ & $6.08\pm0.09$ & $6.58\pm0.14$ \\
AI Scientist & $5.66\pm0.09$ & $6.29\pm0.13$ & $7.53\pm0.07$ & $6.88\pm0.18$ & $6.34\pm0.16$ & $6.16\pm0.16$ \\
AI Scientist-v2 & $6.29\pm0.10$ & $5.39\pm0.17$ & $7.62\pm0.08$ & $7.54\pm0.25$ & $6.53\pm0.14$ & $6.44\pm0.20$ \\
CAR (ours) & $6.89\pm0.08$ & $4.59\pm0.09$ & $6.90\pm0.11$ & $8.80\pm0.14$ & $6.64\pm0.10$ & $6.74\pm0.09$ \\
\bottomrule
\end{tabular}
\end{table}

\begin{table}[t]
\centering\small
\caption{Paired methodological novelty (MN), CAR minus each baseline. For every keyword the difference of the two available-case keyword means is taken (all ideas with a complete panel, without slot matching), so $\Delta$ equals the difference of the absolute MN columns of Table~\ref{tab:main_liveideabench_available} up to rounding. SE over the $40$ keyword differences. $p$: two-sided Wilcoxon signed-rank over keywords, Holm-adjusted within backbone over the four baselines. Ideas: CAR/comparator ideas with a complete panel. Unresolved and insufficient-evidence cases are excluded.}
\label{tab:matched_mn_main}
\begin{tabular}{@{}llrrrrr@{}}
\toprule
Backbone & Comparator & Ideas & CAR MN & Other MN & $\Delta\pm\mathrm{SE}$ & Holm $p$ \\
\midrule
GPT-5.4 & Base & 197/199 & 6.63 & 5.36 & $+1.27\pm0.17$ & $<.001$ \\
GPT-5.4 & Research Ideation Agent & 197/198 & 6.63 & 6.88 & $-0.25\pm0.15$ & $.211$ \\
GPT-5.4 & AI Scientist & 197/197 & 6.63 & 5.81 & $+0.83\pm0.25$ & $.005$ \\
GPT-5.4 & AI Scientist-v2 & 197/197 & 6.63 & 6.23 & $+0.40\pm0.19$ & $.058$ \\
GLM-5.2 & Base & 195/195 & 6.74 & 4.87 & $+1.86\pm0.21$ & $<.001$ \\
GLM-5.2 & Research Ideation Agent & 195/199 & 6.74 & 6.58 & $+0.16\pm0.17$ & $.558$ \\
GLM-5.2 & AI Scientist & 195/197 & 6.74 & 6.16 & $+0.58\pm0.16$ & $.001$ \\
GLM-5.2 & AI Scientist-v2 & 195/151 & 6.74 & 6.44 & $+0.30\pm0.21$ & $.558$ \\
\bottomrule
\end{tabular}
\end{table}

\begin{table}[t]
\centering
\small
\setlength{\tabcolsep}{4pt}
\caption{Full LiveIdeaBench results for the ablation conditions
($40$ keywords $\times$ $5$ ideas per backbone).
O = originality, F = feasibility, C = clarity, Flu. = fluency.
These dimensions report mean $\pm$ SE across keywords.
Flex. is the 30th percentile of keyword means of O/F/C/Flu.,
with bootstrap SE over $20{,}000$ resamples.
MN is $10$ minus mean methodological overlap over three judges,
averaged within keyword and then across keywords, with SE over keywords.
MN uses every idea with a complete panel.
The contrasts in Table~\ref{tab:matched_mn_ablations} are differences
of these keyword means.
Missing values are not imputed.
MN coverage is $3{,}928/4{,}000$ ideas, with $6$ insufficient-evidence
exclusions and $66$ technical failures, including $2$ provider refusals.}
\label{tab:ablations_liveideabench_available}
\begin{tabular}{@{}lcccccc@{}}
\toprule
Method & O $\uparrow$ & F $\uparrow$ & C $\uparrow$
& Flu. $\uparrow$ & Flex. $\uparrow$ & MN $\uparrow$ \\
\midrule
\multicolumn{7}{c}{\textbf{GPT-5.4}} \\
\midrule
CAR (ours) & $6.75\pm0.09$ & $4.97\pm0.11$ & $6.80\pm0.13$ & $8.28\pm0.15$ & $6.49\pm0.07$ & $6.63\pm0.13$ \\
Floor & $6.54\pm0.11$ & $5.10\pm0.10$ & $7.01\pm0.10$ & $8.19\pm0.14$ & $6.57\pm0.06$ & $6.23\pm0.14$ \\
Direct & $6.57\pm0.10$ & $5.02\pm0.11$ & $6.93\pm0.12$ & $8.40\pm0.13$ & $6.55\pm0.08$ & $6.33\pm0.10$ \\
Freeform & $6.69\pm0.10$ & $5.08\pm0.10$ & $6.88\pm0.11$ & $8.21\pm0.13$ & $6.52\pm0.08$ & $6.66\pm0.10$ \\
KG relation-label shuffle & $6.67\pm0.08$ & $5.06\pm0.11$ & $6.90\pm0.11$ & $8.37\pm0.13$ & $6.53\pm0.05$ & $6.67\pm0.09$ \\
No operator validation & $6.56\pm0.09$ & $5.05\pm0.10$ & $6.77\pm0.13$ & $8.36\pm0.12$ & $6.49\pm0.10$ & $6.68\pm0.11$ \\
No \textit{bisociate} & $6.55\pm0.10$ & $5.23\pm0.11$ & $6.99\pm0.12$ & $8.40\pm0.15$ & $6.54\pm0.07$ & $6.44\pm0.12$ \\
No \textit{analogize} & $6.64\pm0.09$ & $4.93\pm0.10$ & $6.79\pm0.11$ & $8.07\pm0.17$ & $6.40\pm0.15$ & $6.69\pm0.10$ \\
No \textit{explore} & $6.63\pm0.10$ & $5.02\pm0.09$ & $6.93\pm0.10$ & $8.37\pm0.13$ & $6.53\pm0.06$ & $6.61\pm0.11$ \\
No \textit{elaborate} & $6.57\pm0.09$ & $4.98\pm0.11$ & $6.81\pm0.13$ & $8.12\pm0.14$ & $6.41\pm0.07$ & $6.65\pm0.09$ \\
\midrule
\multicolumn{7}{c}{\textbf{GLM-5.2}} \\
\midrule
CAR (ours) & $6.89\pm0.08$ & $4.59\pm0.09$ & $6.90\pm0.11$ & $8.80\pm0.14$ & $6.64\pm0.10$ & $6.74\pm0.09$ \\
Floor & $6.81\pm0.08$ & $4.77\pm0.11$ & $7.06\pm0.10$ & $9.05\pm0.09$ & $6.75\pm0.08$ & $6.38\pm0.13$ \\
Direct & $6.83\pm0.08$ & $4.68\pm0.09$ & $7.09\pm0.09$ & $8.94\pm0.11$ & $6.74\pm0.06$ & $6.53\pm0.11$ \\
Freeform & $6.90\pm0.09$ & $4.67\pm0.11$ & $6.95\pm0.10$ & $8.87\pm0.12$ & $6.67\pm0.10$ & $6.36\pm0.13$ \\
KG relation-label shuffle & $6.73\pm0.08$ & $4.68\pm0.10$ & $6.95\pm0.09$ & $9.04\pm0.10$ & $6.71\pm0.05$ & $6.55\pm0.14$ \\
No operator validation & $6.80\pm0.09$ & $4.81\pm0.12$ & $6.97\pm0.11$ & $8.88\pm0.09$ & $6.80\pm0.09$ & $6.64\pm0.11$ \\
No \textit{bisociate} & $6.72\pm0.08$ & $4.72\pm0.11$ & $6.97\pm0.11$ & $9.11\pm0.09$ & $6.74\pm0.05$ & $6.63\pm0.12$ \\
No \textit{analogize} & $6.77\pm0.09$ & $4.57\pm0.10$ & $6.80\pm0.09$ & $8.95\pm0.10$ & $6.58\pm0.10$ & $6.71\pm0.11$ \\
No \textit{explore} & $6.70\pm0.09$ & $4.61\pm0.12$ & $6.91\pm0.10$ & $9.05\pm0.11$ & $6.67\pm0.07$ & $6.39\pm0.15$ \\
No \textit{elaborate} & $6.84\pm0.07$ & $4.46\pm0.09$ & $6.84\pm0.10$ & $8.82\pm0.13$ & $6.62\pm0.06$ & $6.43\pm0.09$ \\
\bottomrule
\end{tabular}
\end{table}

\begin{table}[t]
\centering
\small
\setlength{\tabcolsep}{4pt}
\caption{Paired methodological novelty (MN), CAR minus each ablation
condition. For each keyword, we take the difference of the two
available-case keyword means using all ideas with a complete panel,
without slot matching. Thus, $\Delta$ equals the difference of the
absolute MN columns in Table~\ref{tab:ablations_liveideabench_available}
up to rounding. SE is over the $40$ keyword differences.
$p$ values are from two-sided Wilcoxon signed-rank tests,
Holm-adjusted within each backbone over the nine conditions.
Ideas gives the number of CAR/comparator ideas with a complete panel.
Unresolved and insufficient-evidence cases are excluded.}
\label{tab:matched_mn_ablations}
\begin{tabular}{@{}lrrrrr@{}}
\toprule
Comparator & Ideas & CAR MN & Other MN
& $\Delta\pm\mathrm{SE}$ & Holm $p$ \\
\midrule
\multicolumn{6}{c}{\textbf{GPT-5.4}} \\
\midrule
Floor & 197/196 & 6.63 & 6.23 & $+0.40\pm0.15$ & $.168$ \\
Direct & 197/193 & 6.63 & 6.33 & $+0.31\pm0.14$ & $.168$ \\
Freeform & 197/196 & 6.63 & 6.66 & $-0.02\pm0.12$ & $1.000$ \\
KG relation-label shuffle & 197/199 & 6.63 & 6.67 & $-0.04\pm0.12$ & $1.000$ \\
No operator validation & 197/193 & 6.63 & 6.68 & $-0.05\pm0.11$ & $1.000$ \\
No \textit{bisociate} & 197/196 & 6.63 & 6.44 & $+0.20\pm0.12$ & $.825$ \\
No \textit{analogize} & 197/194 & 6.63 & 6.69 & $-0.06\pm0.13$ & $1.000$ \\
No \textit{explore} & 197/198 & 6.63 & 6.61 & $+0.03\pm0.13$ & $1.000$ \\
No \textit{elaborate} & 197/195 & 6.63 & 6.65 & $-0.01\pm0.09$ & $1.000$ \\
\midrule
\multicolumn{6}{c}{\textbf{GLM-5.2}} \\
\midrule
Floor & 195/196 & 6.74 & 6.38 & $+0.36\pm0.12$ & $.133$ \\
Direct & 195/199 & 6.74 & 6.53 & $+0.20\pm0.12$ & $.797$ \\
Freeform & 195/195 & 6.74 & 6.36 & $+0.37\pm0.12$ & $.076$ \\
KG relation-label shuffle & 195/199 & 6.74 & 6.55 & $+0.19\pm0.13$ & $1.000$ \\
No operator validation & 195/198 & 6.74 & 6.64 & $+0.10\pm0.10$ & $1.000$ \\
No \textit{bisociate} & 195/198 & 6.74 & 6.63 & $+0.11\pm0.11$ & $1.000$ \\
No \textit{analogize} & 195/196 & 6.74 & 6.71 & $+0.02\pm0.12$ & $1.000$ \\
No \textit{explore} & 195/196 & 6.74 & 6.39 & $+0.34\pm0.14$ & $.133$ \\
No \textit{elaborate} & 195/199 & 6.74 & 6.43 & $+0.30\pm0.10$ & $.081$ \\
\bottomrule
\end{tabular}
\end{table}

On the critic dimensions, Holm correction is applied over
the nine ablation contrasts separately within each backbone
and dimension. Six contrasts (CAR minus condition) are
significant on GPT-5.4. For originality, these are no operator
validation ($+0.19$, $p=.036$), no \textit{elaborate}
($+0.18$, $p=.025$), Direct ($+0.18$, $p=.039$) and
Floor ($+0.21$, $p=.047$). For clarity, they are no
\textit{bisociate} ($-0.19$, $p=.045$) and Floor
($-0.21$, $p=.045$). No contrast is significant on GLM-5.2
(smallest adjusted $p=.146$).

\section{Additional analyses}
\label{app:additional}
\label{app:analysis}

\subsection{Scientific domains}
\label{app:domains}
We group the keywords into four scientific domains, with
$10$ keywords per domain. On both backbones and in every domain,
CAR has higher originality and fluency and lower feasibility
and clarity than The AI Scientist and The AI Scientist-v2.
CAR's MN exceeds Base and The AI Scientist in every domain on
both backbones. These are point-estimate patterns, not claims
that every domain-specific contrast is significant.

Against RIA, the MN difference by domain
(physical / life / computational / social) is
$+0.007/-0.273/+0.067/-0.793$ on GPT-5.4 and
$-0.340/+0.422/-0.102/+0.647$ on GLM-5.2.
Domain contrasts use the same available-case keyword means
as the main tables. To test whether contrasts vary across
domains, we compare the variance of the four domain mean
differences against $9{,}999$ random permutations of the
$40$ keyword differences into groups of ten.
Holm correction covers $20$ heterogeneity tests per backbone,
comprising four comparators and five outcomes.
No MN heterogeneity test reaches raw $p<.05$
(minimum $.084$), and no heterogeneity test survives
Holm correction. With ten keywords per domain, this does
not establish domain invariance.

\subsection{Research proposal type}
\label{app:ideatype}
Methods differ in the kinds of contributions they propose.
We drew $120$ ideas ($24$ per main method, $12$ per backbone,
all from the main-comparison generations) and labeled each
with any of three contribution types: a \textsc{method}
(a technique, system or procedure to be built and applied),
a \textsc{study} (an investigation or analysis whose
contribution is learning what is true) or a \textsc{resource}
(a dataset, measure or benchmark offered as the contribution).

Three annotators labeled the $250$-word abstracts without
method identities. For the $119$ ideas with nonempty label
sets from all three annotators, we retain each label endorsed
by at least two. One CAR idea is excluded from this analysis.
Labels may overlap. Pairwise exact agreement on the full
label set is $0.76$--$0.85$, computed on the items available
to each annotator pair. Cohen's $\kappa$ per label is
$0.68$--$0.84$ for \textsc{method}, $0.65$--$0.73$ for
\textsc{study}, and $0.45$--$0.54$ for \textsc{resource}.

Base and RIA predominantly propose methods:
$21/24$ and $24/24$ ideas carry a \textsc{method} label.
CAR and The AI Scientist-v2 carry this label in $16/23$
and $16/24$ cases, and a \textsc{study} label in eight and
seven cases. The AI Scientist has five \textsc{method},
$12$ \textsc{study} and eight \textsc{resource} labels
among its $24$ ideas. On GPT-5.4, where The AI Scientist is the most feasible and least original method, its $12$ ideas carry three\textsc{method}, three \textsc{study} and six \textsc{resource} labels. These descriptive differences do not establish
how much contribution type explains the trade-off.
RIA and Base both predominantly propose methods, yet RIA
scores higher on originality on both backbones.

A GPT-5.4 tagger given the same rubric exactly matches the
majority label set on $58\%$ of the $119$ items.
It matches all $65$ items on which it returns a single label,
but often adds labels not endorsed by the majority when it
returns multiple labels. Its per-label $\kappa$ against the
majority ranges from $0.51$ to $0.59$.
We report the human labels. Agreement with the tagger is
an additional comparison, not evidence that the labels
are free of annotator bias.

\subsection{Originality--feasibility trade-off}
\label{app:axes}

\subsubsection{Expert-rated proposals}
\label{app:axes_sitradeoff}
\begin{figure}[h]
\centering
\includegraphics[width=0.45\linewidth]{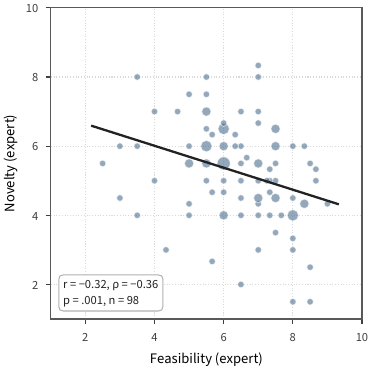}
\caption{Expert novelty and feasibility ratings of
proposals from \citet{si2025can}. Marker area is proportional to the number of proposals with that score pair.}
\label{fig_si_tradeoff}
\end{figure}

Figure~\ref{fig_si_tradeoff} plots the expert novelty
and feasibility ratings of \citet{si2025can}.
Ideas rated as more novel tend to be rated as less
feasible. A similar trade-off therefore appears in expert
ratings on a separate proposal sample, although this does
not rule out bias in our judge panel.

\subsubsection{LiveIdeaBench}
\label{app:axes_liveideabench}
\begin{figure}[h]
\centering
\includegraphics[width=0.95\linewidth]{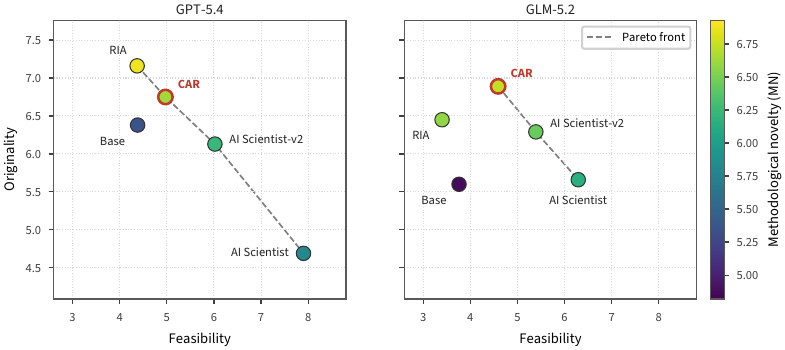}
\caption{Originality against feasibility on LiveIdeaBench. Points are method means over $40$ keywords, and color shows MN. Standard errors over keywords are $0.07$--$0.17$ on both axes and are not drawn. The dashed line is the Pareto front of the five methods.}
\label{fig_liveideabench_pareto}
\end{figure}

Figure~\ref{fig_liveideabench_pareto} plots the method means
of Table~\ref{tab:main}. No method has both the highest
originality and the highest feasibility on either backbone,
so \S\ref{results} considers the two axes together.
CAR lies on the Pareto front of the five methods on both
backbones. On GPT-5.4, the front comprises RIA, CAR,
The AI Scientist-v2 and The AI Scientist. On GLM-5.2, it
comprises CAR, The AI Scientist-v2 and The AI Scientist.
These fronts describe the reported point estimates,
not statistically established dominance.

\subsubsection{Judge-rated proposals}
\begin{figure}[h]
\centering
\includegraphics[width=0.9\linewidth]{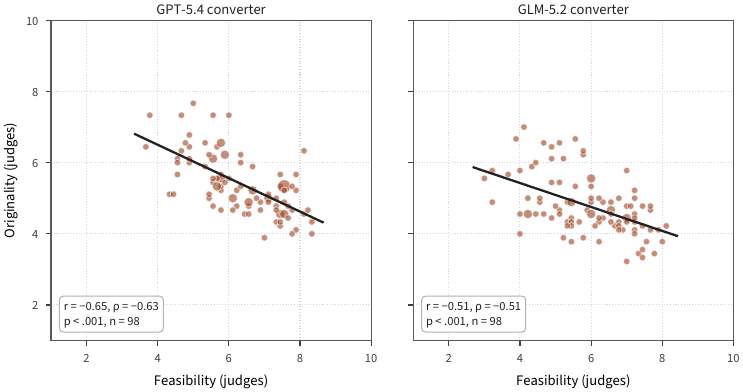}
\caption{Originality and feasibility scores assigned by
our judge panel to individual proposals from \citet{si2025can}. Panel titles give the model that converted each proposal to the 100-word target before judging. Marker area is proportional to the number of proposals with that score pair.}
\label{fig_liveideabench_ovsf}
\end{figure}

Figure~\ref{fig_liveideabench_ovsf} plots originality and
feasibility scores assigned by our judge panel to individual
proposals from \citet{si2025can}. These are the external
validation proposals, not the LiveIdeaBench generations.
The plot describes their score relationship without
establishing its cause.

\subsection{Graph distance (exploratory)}
\label{app:graphdist}
As an exploratory analysis, we relate the graph distance
recorded for CAR's operators to idea scores. Each successful
\texttt{bisociate}, \texttt{explore} and \texttt{analogize}
call stores a hop distance (endpoint span, frontier distance
and analogy reach, respectively), computed by a breadth-first
search capped at six hops. \texttt{Elaborate} records none.
A null value may indicate a distance beyond the cap, a newly
created target, absent graph anchors or disconnected components.
We do not impute these values.

The population is $200$ CAR ideas per backbone, with $197$
GPT-5.4 and $195$ GLM-5.2 complete MN panels.
Eleven GPT-5.4 and $27$ GLM-5.2 ideas have no finite distance.
Nulls account for $117$ of $199$ and $114$ of $202$
\texttt{bisociate} calls, and $125$ of $127$ and $176$ of
$187$ \texttt{analogize} calls, respectively.

With the keyword as the unit ($n=40$), the predictor is the
keyword mean of each idea's maximum recorded finite distance,
excluding ideas with no finite distance. Outcome means use
all valid scored ideas for the keyword.
We use $2{,}000$ whole-keyword bootstrap resamples with
seed $42$ for pointwise $95\%$ confidence intervals.

The Spearman correlation of this predictor with originality
is $-0.44$ for GPT-5.4 (raw $p=.005$, $95\%$ bootstrap CI
$[-0.67,-0.13]$, Holm-adjusted $p=.098$) and $-0.17$ for
GLM-5.2 (raw $p=.29$). For MN, it is $-0.11$ (raw $p=.48$)
and $-0.25$ (raw $p=.12$).
The fraction of unresolved distances is positively associated
with originality for GPT-5.4 ($\rho=.38$, raw $p=.016$,
adjusted $p=.31$) and with MN for GLM-5.2 ($\rho=.38$,
raw $p=.015$, adjusted $p=.36$).
No association survives Holm correction over the $20$
GPT-5.4 and $25$ GLM-5.2 tests.

These are between-keyword associations, subject to domain
and corpus-size confounding. The pooled distance proxy mixes
three operator-specific measures and conditions on finite
distances. These results do not show that graph distance
raises or lowers novelty.

\section{Operator validation and provenance}
\label{app:traces}
\textbf{Failure modes and repairs.}
Table~\ref{tab:rejections} summarizes rejected operator calls
in the CAR arm of the main experiment.
GPT-5.4 has $16$ rejections in $721$ attempts, affecting
$14$ of $200$ ideas. GLM-5.2 has $202$ rejections in
$935$ attempts, affecting $126$ of $200$ ideas.
Most GLM-5.2 rejections concern missing type or description
fields for a new entity.

Of the $218$ rejected calls, $176$ are immediately followed
by an accepted call of the same operator. For $213$, the
next successful operator call uses the same operator,
potentially after further rejections. Several rejected calls
can therefore lead to the same successful call. These counts
measure recovery in the operation sequence, not whether the
accepted call preserves the rejected call's intended meaning.
All $400$ ideas were completed.

For example, a \texttt{bisociate} call proposing
\emph{activation-contingent reserve contracts} as a bridge
was rejected because that entity already existed.
The next call succeeded with the same endpoints, a new bridge
named \emph{disruption-indexed exercise ladder}, and revised
relation labels.

\begin{table}[t]
\centering
\footnotesize
\setlength{\tabcolsep}{4pt}
\caption{Rejected operator calls for CAR ($40$ keywords
$\times$ $5$ ideas per backbone).
B/A/X/E denote \textit{bisociate}, \textit{analogize},
\textit{explore} and \textit{elaborate}.
Accepted calls number $199/127/179/200$ on GPT-5.4
and $202/187/142/202$ on GLM-5.2, totaling $1{,}438$.
The final column describes corrective actions.
Aggregate recovery counts are reported in the text.}
\label{tab:rejections}
\begin{tabular}{@{}lrrrrrrp{4.6cm}@{}}
\toprule
& \multicolumn{2}{c}{GPT-5.4}
& \multicolumn{4}{c}{GLM-5.2} & \\
\cmidrule(lr){2-3}\cmidrule(lr){4-7}
Validator error & B & A & B & A & X & E & Typical repair \\
\midrule
\multicolumn{8}{@{}l}{\emph{Grounding and design constraints}} \\
Endpoint not in graph
& 1 & 1 & 20 & 10 & 5 & 0
& ungrounded endpoint swapped for an existing node \\
Analogy relation mismatch
& 0 & 2 & 0 & 26 & 0 & 0
& target relation rewritten to match the source relation \\
Analogy target shares source endpoint
& 0 & 1 & 0 & 7 & 0 & 0
& disjoint pair chosen \\
Bridge entity already exists
& 7 & 0 & 0 & 0 & 0 & 0
& new bridge concept \\
Path not exactly two edges
& 0 & 0 & 4 & 0 & 0 & 0
& path shortened to source--bridge--target \\
Inconsistent bridge or endpoints
& 0 & 0 & 3 & 0 & 0 & 0
& bridge named consistently \\
No new edges
& 0 & 2 & 0 & 0 & 0 & 2
& same operator, or a move to \texttt{explore} \\
Paper node as operand
& 0 & 0 & 0 & 0 & 0 & 1
& grounded on a non-paper entity \\
\midrule
\multicolumn{8}{@{}l}{\emph{Malformed arguments}} \\
New entity lacks type or description
& 0 & 2 & 90 & 19 & 3 & 0
& missing type or description supplied \\
Missing relation description
& 0 & 0 & 2 & 5 & 0 & 0
& relation description supplied \\
Triplet fields missing
& 0 & 0 & 2 & 3 & 0 & 0
& missing triplet fields supplied \\
\midrule
Total & 8 & 8 & 121 & 70 & 8 & 3 & \\
\bottomrule
\end{tabular}
\end{table}

\textbf{Declared provenance.}
We audit eight method--backbone groups: CAR, Freeform, Direct,
Floor and no operator validation on GPT-5.4, and CAR, Freeform
and Direct on GLM-5.2. These comprise $1{,}600$ ideas and
$16{,}820$ declared entity indices, counted once per idea.
For each idea, we compare the declared indices with indexed
appearances in its log before proposal submission, excluding
tool-call lines from the matching evidence.

All but one declaration, in one GPT-5.4 CAR idea, have a
matching appearance. Coverage is $99.94\%$ for GPT-5.4 CAR
and $100\%$ for the other seven audited groups.
Among CAR's declared entities, $657/1{,}652$ ($39.8\%$)
on GPT-5.4 and $715/1{,}999$ ($35.8\%$) on GLM-5.2
also match literature entities named in accepted operations.
Every CAR idea declares at least one such entity.
This is an audit of recorded index appearances and operation
arguments. It does not establish that a declared entity was
necessary to the idea or that an operation was scientifically
sound.

\section{Case studies}
\label{app:cases}
We examine two illustrative cases: a proposed mechanism
implemented in the executor's code and a proposed mechanism
not identified in the saved implementation.
These examples do not estimate how often the executor
preserves CAR's proposed mechanisms.

\subsection{From trace to code: pan-cancer survival}
\label{app:case_pancancer}
On BioXArena's pan-cancer survival task, a
\textit{bisociate} operation connects
\emph{Guided Mixture of Experts} with a discrete-time
survival loss through \emph{interval-specialized expert routing}.
The code snapshot with the highest validation score
implements a mixture of experts with a gate conditioned
on cancer type and survival interval.
Specifically, the gate takes concatenated cancer and
interval embeddings and gene-availability indicators,
then assigns softmax weights to the expert outputs.

The reported CAR test c-index on this task is $0.775$,
compared with $0.662$ for the same executor using its own
ideas and $0.785$ for the human reference.
The inspected code provides an example of the proposed
routing mechanism appearing in an implementation.
The task-level comparison does not isolate the contribution
of that mechanism to the score.

\subsection{An omitted training-time mechanism: OpenOOD}
\label{app:case_openood}
On the OpenOOD task of FML-bench, CAR proposes an
entropy-gated feature-mixup objective applied during training
(\texttt{egfm\_npace}). The task description invites changes
to the training procedure, but the editable files are
\texttt{algorithm.py} and \texttt{split\_config.json}.
These expose the model, training hyperparameters and
post-hoc scoring function, while the standard cross-entropy
loss is defined in a training script outside the editable set.

A lexical inspection of the $100$ saved code snapshots
found no explicit feature-mixup implementation.
The run reaches validation area under the receiver operating characteristic curve (AUROC) $0.914$, against a baseline
of $0.876$, and test AUROC $0.981$.
These scores do not establish the effectiveness of the
proposed training-time mechanism, whose implementation
was not identified in the saved code.
The logs do not record why the executor omitted it.
The restricted interface is relevant context, but does
not establish that implementation was impossible.

\section{Computational cost}
\label{app:compute}

\subsection{LiveIdeaBench}
Table~\ref{tab:cost} reports recorded generation-cost estimates
per saved idea. On GPT-5.4, CAR costs about four times as much
as The AI Scientist variants and is comparable to RIA.
The graph explorer accounts for about $70\%$ of CAR's tokens
but $12\%$ of its estimated cost, reflecting its use of a
cheaper model. About $87\%$ of CAR's input tokens are
recorded as cache reads.

Freeform is slightly more expensive than CAR on GPT-5.4
(\$0.193 versus \$0.185 per idea), but cheaper on GLM-5.2
(\$0.032 versus \$0.039). Direct and Floor cost approximately
$35$--$37\%$ less than CAR on GPT-5.4 and $26$--$27\%$ less
on GLM-5.2. These comparisons exclude graph construction,
which is shared across ideas and graph-based conditions.

The estimates cover logged calls associated with saved ideas,
including tracked subcalls and retries. They do not provide
complete accounting of failed or unsaved attempts.
In particular, costs for the $47$ missing GLM-5.2
AI Scientist-v2 outputs are not included.

\begin{table}[t]
\centering
\small
\setlength{\tabcolsep}{4pt}
\caption{Estimated idea-generation cost per idea on LiveIdeaBench.
Costs use logged tokens and list prices with cached-input discounts,
rather than provider receipts. Input tokens include cached tokens.
Token columns report means per saved idea.
Graph construction, text normalization and scoring are excluded.
Base generations did not log usage.
$^\dagger$The AI Scientist-v2 saved $153$ of $200$ requested ideas
on GLM-5.2. Its row summarizes those saved ideas.}
\label{tab:cost}
\begin{tabular}{@{}lrrrrr@{}}
\toprule
Method & $n$ & USD mean & USD median & In tok. & Out tok. \\
\midrule
\multicolumn{6}{c}{\textbf{GPT-5.4}} \\
\midrule
RIA & 200 & 0.160 & 0.146 & 6.9k & 9.5k \\
AI Scientist & 200 & 0.046 & 0.046 & 7.0k & 2.0k \\
AI Scientist-v2 & 200 & 0.046 & 0.044 & 11.2k & 1.7k \\
CAR & 200 & 0.185 & 0.185 & 268.0k & 17.1k \\
\quad idea agent / explorer & 200 & 0.163 / 0.022 & -- & 81.2k / 186.8k & 2.9k / 14.2k \\
Floor & 200 & 0.117 & 0.115 & 213.3k & 12.7k \\
Direct & 200 & 0.121 & 0.119 & 223.6k & 13.3k \\
Freeform & 200 & 0.193 & 0.191 & 260.9k & 13.8k \\
\midrule
\multicolumn{6}{c}{\textbf{GLM-5.2}} \\
\midrule
RIA & 200 & 0.0069 & 0.0059 & 7.4k & 6.1k \\
AI Scientist & 200 & 0.0056 & 0.0051 & 5.8k & 5.4k \\
AI Scientist-v2 & 153$^\dagger$ & 0.0070 & 0.0050 & 15.6k & 5.3k \\
CAR & 200 & 0.039 & 0.037 & 322.7k & 19.1k \\
\quad idea agent / explorer & 200 & 0.026 / 0.014 & -- & 209.5k / 113.2k & 11.6k / 7.5k \\
Floor & 200 & 0.029 & 0.028 & 192.8k & 15.7k \\
Direct & 200 & 0.029 & 0.028 & 194.6k & 16.0k \\
Freeform & 200 & 0.032 & 0.030 & 240.3k & 16.2k \\
\bottomrule
\end{tabular}
\end{table}

\subsection{FML-bench}
Across the $24$ CAR runs on the eight Lite tasks with three
seeds, a $100$-step run took a median of $35$ hours
(range $6.5$--$63$ hours, $803$ hours in total).
The median executor usage was $1.35$M GPT-5.4 tokens
(range $0.77$--$4.73$M).
Runs used one NVIDIA A100 (40\,GB) each.
These figures come from the runs' \texttt{summary.json}
records and describe elapsed execution time, not measured
GPU-active time. Training, agent calls and other execution
overheads contribute to this duration.
The per-idea LiveIdeaBench costs in Table~\ref{tab:cost}
are not measurements of downstream idea-generation cost.

\subsection{BioXArena}
Runs were allocated a nominal two-hour budget including
idea generation, with final evaluation performed separately.
Across the $32$ recorded CAR runs on the ten pilot tasks,
elapsed execution time ranges from $1.8$ to $2.3$ hours.
The executor completed a median of $30$ search steps
(range $13$--$103$) and used a median of $0.60$M GPT-5.4
tokens (range $0.17$--$1.62$M).
Runs used one A100 (40\,GB) each.

This accounting includes preliminary runs as well as the
final graded runs, so it does not describe only the runs
underlying the reported benchmark table.
The nominal budget should not be interpreted as a strict
cap on total elapsed time.

\section{Limitations}
\label{app:limitations}

\textbf{Literature and extraction.}
CAR builds its graph from papers whose full text it can obtain.
In our experiments, some papers that could not be downloaded
automatically were retrieved manually through institutional
subscriptions. Deployments without equivalent access may
construct different or less extensive corpora.
Entities and relations are extracted by an LLM, and we do not
measure extraction accuracy against expert annotation.
Relation repair addresses missing connectivity but does not
verify scientific correctness. Name-based merging can leave
semantically overlapping entities separate.

\textbf{Cost.}
Graph construction and iterative ideation introduce additional
calls and token usage. Recorded generation costs vary by
backbone and method (Appendix~\ref{app:compute}).
Building the graph once per topic amortizes construction
across ideas, but not the per-idea reasoning and retrieval
costs. The reported estimates exclude evaluation and
unrecorded failed attempts, so they are not complete
end-to-end costs.

\textbf{Operators and auditability.}
The ablations do not establish a consistent quality benefit
from explicit operators beyond free-form iteration.
Some operator and validation contrasts are significant on
one backbone, so these results should not be interpreted
as showing that the components have no effect.
The trace records tool interactions, accepted and rejected
operations, and declared literature entities.
Its guarantees are procedural and structural.
An entity appearing in a tool result does not establish
that the agent used it meaningfully, and an accepted
operation need not be scientifically sound.
The trace does not establish that each entity or operation
was causally necessary to the proposal.

\textbf{Evaluation.}
Idea quality is assessed by LLM judges, without direct human
evaluation of the generated ideas' quality.
Agreement with expert ratings is assessed on a separate
sample of NLP proposals. This sample also informed development
of the methodological novelty evaluator, so its reported
agreement is not independent held-out validation.
The adjusted split-half human-agreement estimate is a
reference, not a second observed expert panel, and the
comparison does not establish equivalence to human reviewers.
Retrieval-based novelty is conditional on the evidence found
and cannot establish novelty across the wider literature.

The $40$-keyword sample limits precision, especially for
domain analyses with ten keywords per domain.
Results differ between backbones, and nonsignificant
contrasts do not establish the absence of an effect.
Missing generations and incomplete MN panels yield
available-case comparisons whose exclusions may be selective
(Appendix~\ref{app:coverage}).
Execution evaluates CAR's proposals through one executor
on machine learning tasks, with a single run per task on
BioXArena. Performance depends on the executor's implementation
and the task interface as well as the proposal.
These experiments do not establish effectiveness across
scientific research more broadly.

\section{Prompts}
\label{app:prompts}

\subsection{Knowledge construction}
\subsubsection{Literature search}
\label{app:prompts_knowledge_search}

The following literature search prompts are adapted from \citet{si2025can}. The first-round prompt below applies to task-based search.
In topic mode, used for LiveIdeaBench, the implementation
instead submits the lowercased, trimmed topic directly as
\texttt{KeywordQuery}, without an LLM call.
Subsequent rounds use the second prompt.

\begin{llmprompt}{Literature Search Prompt: First Round}
\begin{Verbatim}[breaklines=true, breaksymbolleft=]
You are a researcher doing literature review on this research task:
{research_task}

You should propose some keywords for querying the Semantic Scholar API to find the most relevant papers to the research task. You can start by extracting and combining the keywords from the research task.

Formulate your query as: KeywordQuery("query"). Just give me one query, and the query can be a concatenation of multiple keywords (just put a space between every word); but please be concise and try to cover the most important aspects.

Your query (just return the query itself with no additional text):
\end{Verbatim}
\end{llmprompt}

\begin{llmprompt}{Literature Search Prompt: Subsequent Rounds}
\begin{Verbatim}[breaklines=true, breaksymbolleft=]
You are a researcher. You are doing literature review on this research task:
{research_task}

You should propose some queries for using the Semantic Scholar API to find the most relevant papers to this research task.

You are allowed to use the following functions for making queries:
(1) KeywordQuery("keyword"): find most relevant papers to the given keyword (the keyword shouldn't be too long and specific, otherwise the search engine will fail; it is ok to combine a few short keywords with spaces, such as "language model reasoning").
(2) PaperQuery("paperId"): find the most similar papers to the given paper (as specified by the paperId).
(3) GetReferences("paperId"): get the list of papers referenced in the given paper (as specified by the paperId).

Right now you have already collected the following relevant papers:
{grounding_papers_str}

You can formulate new search queries based on these papers. And you have already asked the following queries:
{queries_str}

Please formulate a new query to expand our paper collection with more diverse and relevant papers (you can do so by diversifying the types of queries and minimizing the overlap with previous queries; e.g., try to include all three types of queries if possible). Directly give me your new query without any explanation or additional text, just the query itself:
\end{Verbatim}
\end{llmprompt}

For task-based search, \texttt{\{research\_header\}} contains
``Research task:'' followed by the task description and
a blank line. The relevance criterion is:

\begin{llmprompt}{Paper Relevance Criterion: Task Mode}
\begin{Verbatim}[breaklines=true, breaksymbolleft=]
(1) The paper is directly relevant to the research task. Note that it should be specific to solve the problem of focus, rather than just generic methods.
\end{Verbatim}
\end{llmprompt}

For topic mode, the header is empty and the relevance
criterion is:

\begin{llmprompt}{Paper Relevance Criterion: Topic Mode}
\begin{Verbatim}[breaklines=true, breaksymbolleft=]
(1) The paper is directly relevant to the topic of: {research_task}. Note that it should be specific to solve the problem of focus, rather than just generic methods.
\end{Verbatim}
\end{llmprompt}

The appropriate header and criterion are inserted into
the following prompt.

\begin{llmprompt}{Paper Relevance Scoring Prompt}
\begin{Verbatim}[breaklines=true, breaksymbolleft=]
{research_header}You are a helpful literature review assistant whose job is to read the below set of papers and score each paper. The criteria for scoring are:
{relevance_criterion}
(2) The paper is an empirical paper that proposes a novel method and conducts empirical experiments to show improvement over baselines (position or opinion papers, review or survey papers, and analysis papers should get low scores for this purpose).
(3) The paper is interesting, exciting, and meaningful, with potential to inspire many new projects.

The papers are:
{formatted_papers}

Please score each paper from 1 to 10. Write the response in JSON format with "paperID: score" as the key and value for each paper.
\end{Verbatim}
\end{llmprompt}

\subsubsection{Knowledge extraction}
\label{app:prompts_knowledge_extraction}
\begin{llmprompt}{Knowledge Extraction Prompt}
\begin{Verbatim}[breaklines=true, breaksymbolleft=]
I want to extract structured, detailed scientific knowledge from the full text of a scientific paper.

Your output will populate a knowledge graph with scientific units, concepts, and relations. Extract the technically important scientific units, concepts, and relations with high coverage, while avoiding generic background mentions that do not help connect the graph.

WORKFLOW:
1. Scientific Units Extraction:
    - Paper: A dense, technical summary of the paper's core contributions (2-3 paragraphs). This paper node will act as the anchor for knowledge extracted from this paper. The graph builder will connect it to the paper's extracted problems, methods, experiments, results, claims, limitations, and formulas, while entity-entity relations may also connect those nodes directly.
    - Problems: Specific challenges, limitations of prior work, or research gaps addressed.
    - Methods: Exact names of novel algorithms, architectures, protocols, or methodologies proposed or substantially modified by this paper. Do not put baselines, prior work, off-the-shelf architectures, standard preprocessing steps, optimizers, or calibration procedures here; put them in the "concepts" list with type "method".
        - Put a technique in top-level "methods" only if the paper proposes it, substantially modifies it, or treats it as a named component of its contribution. Put prior architectures, baselines, standard preprocessing steps, optimizers, metrics, datasets, and off-the-shelf calibration procedures in "concepts" with the most specific type.
    - Experiments: Specific evaluation protocols, benchmark settings, ablation studies, simulation studies, lab assays, user studies, or analysis procedures conducted by the paper. An experiment is the setup/procedure, not the measured outcome.
    - Results: Concrete empirical findings or observations, preferably with metrics, datasets, conditions, confidence intervals, comparisons, or qualitative observations. A result is evidence, not the broad claim it supports.
    - Claims: Declarative scientific assertions made by the paper, usually supported by results. A claim states what the authors argue is true; it is not a raw metric result.
    - Limitations: Explicitly stated weaknesses, failure modes, assumptions, boundary conditions, threats to validity, negative findings, or future-work gaps.
2. Mathematical Formulas Extraction:
    - Extract crucial formulas (in LaTeX) that define the methods or problems.
    - Provide a summary and TeX for each formula.
3. Scientific Concepts Extraction:
    - Method: A prior algorithm, baseline model, experimental protocol, or technique (e.g., "diffusion model", "crispr-cas9", "x-ray crystallography").
    - Metric: A quantitative measure of performance or property (e.g., "fid score", "binding affinity", "tensile strength").
    - Task: The specific application or problem being addressed (e.g., "image generation", "protein folding", "alloy synthesis").
    - Material: A dataset, software library, biological sample, or physical resource (e.g., "cifar-10", "escherichia coli", "graphene").
    - Assumption: Theoretical preconditions, data assumptions, or simplifications (e.g., "i.i.d. data", "ideal gas", "incompressible fluid").
    - Hyperparameter: Key implementation parameters or constants (e.g., "learning rate", "annealing temperature", "catalyst concentration").
    - Implementation Detail: Specific engineering choices or procedural tweaks (e.g., "layer normalization", "centrifugation speed", "quenching rate").
    - Concept: General scientific terms, mechanisms, artifacts, failure modes, or theoretical constructs that don't fit the above (e.g., "entropy", "homeostasis", "manifold")
4. Scientific Relations Extraction: 
    - IMPORTANT: Link the Problems, Methods, Experiments, Results, Claims, Limitations, Concepts, and Formulas to one another using relations.
    - Relation coverage requirement: every important non-paper scientific unit or concept should participate in at least one non-paper relation. Prefer unit-concept, unit-unit, and concept-concept edges over paper-to-concept scaffolding.
    - If a formula is important enough to extract, include at least one explicit relation connecting the formula to the method, metric, problem, concept, experiment, result, claim, or limitation it defines, formalizes, models, computes, optimizes, constrains, or evaluates.
    - Prioritize relations that connect methods to problems, experiments to methods/datasets/metrics, results to experiments/claims, claims to evidence, limitations to affected methods/settings, and concepts to the mechanisms or assumptions they influence.
    - Define the relation name precisely and concisely, and provide a detailed relation description. Explain how and why they are related in detail.
    - Account for 6 types of relations when they are supported by the paper:
        taxonomic (e.g., "is variant of", "is subtype of", "generalizes"),
        compositional (e.g., "contains component", "combines with"),
        mechanistic (e.g., "causes", "regularizes", "constrains", "formalizes", "calibrates", "inhibits", "stabilizes", "modulates"),
        methodological (e.g., "trains on", "optimizes", "evaluates on", "defines metric", "applies to"),
        evidential (e.g., "supports claim", "contradicts claim", "qualifies claim"),
        and evaluative (e.g., "outperforms", "improves metric", "fails under", "compares against").
      Use the domain-appropriate label that most precisely describes the scientific relationship. These examples are illustrative, not exhaustive — use domain-native terms (e.g., "degrades", "binds to", "encodes") when they are more accurate.
    - Avoid overly generic labels such as "uses", "solves", "produces", "based on", or "related to" when a more specific label exists.
    - For example, link a Method to the Problem it addresses, or a Result to the Method that achieved it via "outperforms" or "improves metric".

GENERAL GUIDELINES:
- Return only a valid JSON object. Do not include markdown fences, comments, prose, trailing commas, or ellipses.
- Use exactly these top-level keys: "paper", "problems", "methods", "experiments", "results", "claims", "limitations", "formulas", "concepts", "relations".
- All top-level list fields must be arrays. If nothing is found for a field, return [].
- Be thorough but selective. Aim for high coverage of technically important information, not exhaustive extraction of every mentioned term.
- Extract enough concepts and relations to represent the paper's scientific structure.
- Extract enough non-paper relations for the graph to be dominated by scientific structure rather than document provenance.
- Avoid isolated concepts. Only extract a standalone concept if it has a specific technical summary and either participates in at least one non-paper relation or is necessary context for understanding a formula, method, experiment, result, claim, or limitation.
- Do not extract generic background concepts merely because they are mentioned.
- Ensure all extracted scientific knowledge is accurately represented.
- Provide accurate and technical details in the entity summaries and relation descriptions.
- Ensure mathematical precision for formulas (use LaTeX if needed).
- Entity and relation names must be concise, atomic phrases (1-5 words) and in lowercase to prevent duplicates. Do NOT use full sentences as entity names. Put detailed explanations in the 'summary' field.
- Do not duplicate the same scientific object across multiple top-level lists. Use the most specific unit type: proposed contributions go in methods, prior/background methods go in concepts, empirical procedures go in experiments, measurements go in results, and interpretations go in claims.
- Ensure all keys are present even if lists are empty.
- Extract specific values for hyperparameters and specific implementation choices as context for relations.
- Before returning, internally check that proposed methods are linked to the problems, claims, experiments, or limitations they address; experiments are linked to the methods, datasets/materials, metrics, and results they evaluate; results are linked to the experiments that produced them and the claims they support or qualify; claims are linked to supporting results or reasoning; limitations are linked to affected methods, experiments, datasets/materials, assumptions, or claims; important concepts are linked through supported taxonomic, compositional, mechanistic, methodological, evidential, or evaluative relations; and generic concepts that remain isolated are removed.

I have retrieved the following content from the paper '{paper title}': {content}

Please return a valid JSON object with the following structure. Each array may contain zero, one, or many objects of the shown form. The single object shown in each array is only a template; include as many objects as are supported by the paper.
{
    "paper": {
        "name": "{paper title}",
        "summary": "detailed summary text"
    },
    "problems": [
        {
            "name": "specific problem statement",
            "summary": "detailed problem description (MUST include specific context, datasets, values, or conditional parameters)",
            "relation_description": "edge-specific explanation of how the paper frames, motivates, or operationalizes this problem, including the task setting, failure of prior work, or research gap when available"
        }
    ],
    "methods": [
        {
            "name": "method name",
            "summary": "detailed method description (MUST include specific context, datasets, values, or conditional parameters)",
            "relation_description": "edge-specific explanation of what the paper newly proposes or modifies in this method, and which problem, claim, experiment, or limitation it is meant to address"
        }
    ],
    "experiments": [
        {
            "name": "specific experiment, evaluation, ablation, assay, simulation, or analysis name",
            "summary": "detailed experiment setup/procedure description (MUST include specific datasets, baselines, metrics, controls, conditions, sample sizes, or implementation settings when available)",
            "relation_description": "edge-specific explanation of the setup/procedure run by the paper and what method, dataset, metric, claim, or hypothesis it evaluates"
        }
    ],
    "results": [
        {
            "name": "quantitative or qualitative result",
            "summary": "detailed result description (MUST include specific context, datasets, values, or conditional parameters)",
            "relation_description": "edge-specific explanation of the measurement or observation reported, the condition under which it was obtained, and what claim, method, or comparison it supports or contradicts"
        }
    ],
    "claims": [
        {
            "name": "declarative scientific claim",
            "summary": "detailed claim description, including what the paper asserts and the conditions or scope under which the claim is meant to hold",
            "relation_description": "edge-specific explanation of the evidence, reasoning, experiment, or result in the paper that supports this assertion"
        }
    ],
    "limitations": [
        {
            "name": "specific limitation, failure mode, assumption, or open gap",
            "summary": "detailed limitation description, including affected methods, datasets, settings, assumptions, or boundary conditions when available",
            "relation_description": "edge-specific explanation of the assumption, failure case, missing evidence, scope boundary, or unresolved gap that makes this a limitation"
        }
    ],
    "formulas": [
        {
            "name": "formula name or label",
            "summary": "what does this formula represent",
            "tex": "latex string"
        }
    ],
    "concepts": [
        {
            "name": "concept name",
            "type": "method|metric|task|material|assumption|
            hyperparameter|implementation_detail|concept",
            "summary": "detailed definition, value, or description of the concept (MUST include specific context, datasets, values, or conditional parameters)"
        }
    ],
    "relations": [
        {
            "subject": "entity name",
            "object": "entity name",
            "relation": "concise, domain-specific relation label (not generic like 'uses' or 'related to')",
            "description": "edge-specific technical explanation of exactly how and why the subject relates to the object in this paper (MUST include specific context, datasets, values, or conditional parameters)"
        }
    ]
}
\end{Verbatim}
\end{llmprompt}

\subsubsection{Relation coverage repair}
\label{app:prompts_knowledge_repair}
\begin{llmprompt}{Relation Coverage Repair Prompt}
\begin{Verbatim}[breaklines=true, breaksymbolleft=]
I need to repair relation coverage for one extracted scientific knowledge graph.

Paper: {paper_title}

The extraction already has good entities, but some non-paper nodes are isolated from non-paper scientific relations. Add only relations that are directly supported by the paper text implied by the entity summaries. Do not add new entities. Do not repeat existing relations.

All available entities:
{entity_lines}

Currently isolated nodes that need relation coverage:
{isolated_lines}

Existing relations:
{existing_relation_lines}

Return only valid JSON with this structure:
{
  "relations": [
    {
      "subject": "existing entity name",
      "object": "existing entity name",
      "relation": "short precise relation phrase",
      "description": "specific explanation of how and why the relation holds in this paper"
    }
  ]
}

Requirements:
- Use only exact entity names from the available entity list. Do not append the entity type as a parenthetical suffix (e.g., write "X" not "X (claim)" or "X (formula)").
- Prefer relations involving the isolated nodes.
- Use concise, domain-specific relation labels. Choose from taxonomic (e.g., "is variant of"), compositional (e.g., "contains component"), mechanistic (e.g., "causes", "constrains", "inhibits"), methodological (e.g., "trains on", "evaluates on", "defines metric"), evidential (e.g., "supports claim", "contradicts claim"), or evaluative (e.g., "outperforms", "fails under") types. Use domain-native terms when more accurate.
- Do not use vague labels such as "uses", "solves", "produces", "based on", or "related to".
- Return [] if no supported relation can be added.
\end{Verbatim}
\end{llmprompt}

\subsection{Research ideation}
\label{app:prompts_ideagen}
\subsubsection{Base LLM}
We use LiveIdeaBench's \citep{ruan2026evaluating} original idea generation prompt.
\begin{llmprompt}{Base LLM: Idea Generation Prompt}
\begin{Verbatim}[breaklines=true, breaksymbolleft=]
I'll be submitting your next responses to a "Good Scientific Idea" expert review panel. If they consider your idea to be a good one, you'll receive a reward. Your assigned keyword is: "{keywords}". You may provide background information. The idea MUST be concisely expressed within 100 words total (including any background information). (Note: good scientific ideas should be original (novel contribution), feasible (technically implementable), clearly articulated, and address meaningful problems in the field.).
\end{Verbatim}
\end{llmprompt}

\subsubsection{Research Ideation Agent}
We use Research Ideation Agent's \citep{si2025can} released prompts, with the natural-language-processing qualifiers removed so that they apply to every domain in the benchmark: ``expert researcher in AI'' becomes ``expert researcher'', ``top AI conferences like ACL and NeurIPS'' becomes ``top conferences in the field'', ``large language models'' becomes ``current methods'', and the sentences that restrict ideas to prompting or finetuning are replaced by ``Focus on proposing novel empirical methods''. Everything else, including the idea format, the few-shot examples and the proposal-expansion prompt, is as released. The displayed generation prompt is the retrieval-augmented
variant. Plain-generation batches omit the paper block and
the two sentences beginning ``The above papers'' and
``Instead, you should make sure''.
The instruction to avoid existing ideas is included only
when earlier ideas are available.
\begin{llmprompt}{Research Ideation Agent: Idea Generation Prompt}
\begin{Verbatim}[breaklines=true, breaksymbolleft=]
You are an expert researcher. Now I want you to help me brainstorm some new research project ideas on the topic of: {task_description}.

Here are some relevant papers on this topic just for your background knowledge:
{grounding_papers}
You should generate {ideas_n} different ideas on this topic. Try to be creative and diverse in the idea generation, and do not repeat any similar ideas. The above papers are only for inspiration and you should not cite them and just make some incremental modifications. Instead, you should make sure your ideas are novel and distinct from the prior literature. You should aim for projects that can potentially win best paper awards at top conferences in the field.
Each idea should be described as: (1) Problem: State the problem statement, which should be closely related to the topic description and something that current methods cannot solve well yet. (2) Existing Methods: Mention some existing benchmarks and baseline methods if there are any. (3) Motivation: Explain the inspiration of the proposed method and why it would work well. (4) Proposed Method: Propose your new method and describe it in detail. The proposed method should be maximally different from all existing work and baselines, and be more advanced and effective than the baselines. You should be as creative as possible in proposing new methods, we love unhinged ideas that sound crazy. This should be the most detailed section of the proposal. (5) Experiment Plan: Specify the experiment steps, baselines, and evaluation metrics.
You can follow these examples to get a sense of how the ideas should be formatted (but don't borrow the ideas themselves):
{examples}
You should make sure to come up with your own novel and different ideas for the specified problem: {task_description}. You should try to tackle important problems that are well recognized in the field and considered challenging for current methods. For example, think of novel solutions for problems with existing benchmarks and baselines. In rare cases, you can propose to tackle a new problem, but you will have to justify why it is important and how to set up proper evaluation.
Focus on proposing novel empirical methods. The proposed method section should specify all the details involved, such as how to get the data, etc.
You should avoid repeating the following existing ideas and try to be different and diverse: {existing_ideas}
Please write down your {ideas_n} ideas (each idea should be described as one paragraph. Output the ideas in json format as a dictionary, where you should generate a short idea name (e.g., "Non-Linear Story Understanding", or "Multi-Agent Negotiation") as the key and the actual idea description as the value (following the above format). Do not repeat idea names or contents.
\end{Verbatim}
\end{llmprompt}
\begin{llmprompt}{Research Ideation Agent: Proposal Expansion Prompt}
\begin{Verbatim}[breaklines=true, breaksymbolleft=]
You are an expert researcher and your job is to expand a brief project idea into a full project proposal with detailed experiment plans so that your students can follow the steps and execute the full project. I will provide you with an idea on the topic of: {task_description}.

The idea is:
{idea}
Now you should come up with the full experiment plan covering:
1. Title: A concise statement of the main research question to be used as the paper title.
2. Problem Statement: Clearly define the problem your research intends to address. Explain clearly why this problem is interesting and important.
3. Motivation: Explain why existing methods (both classic ones and recent ones) are not good enough to solve the problem, and explain the inspiration behind the new proposed method. You should also motivate why the proposed method would work better than existing baselines on the problem.
4. Proposed Method: Explain how the proposed method works, describe all the steps. Make sure every step is clearly described and feasible to implement.
5. Step-by-Step Experiment Plan: Break down every single step of the experiments, make sure every step is executable. Cover all essential details such as the datasets, models, and metrics to be used. If the project involves prompting, give example prompts for each step. If the project involves training, describe the training data, objective, and evaluation metrics.
6. Test Case Examples: Give two concrete examples. The first example should show how the baseline method fails on the test case. If there are multiple baselines, give examples for all of them. The second example should show how the proposed method succeeds on the test case. For each test case, include the input and the expected output. You should also provide an explanation for why the outputs from the proposed prompt are better. If the proposed method has multiple steps, break them down into intermediate steps.
7. Fallback Plan: Propose some alternative plans for what should the students do if the proposed method didn't manage to satisfy the success criteria. For example, you can suggest additional analysis to help debug why the proposed method didn't work, which could inform alternative new methods, or turn the project into an analysis paper instead by offering some interesting ablation and insights. Write a coherent paragraph rather than a list of bullet points.
The experiment plan should not include any background introduction (you can skip the literature review, paper writing tips, and ethical discussion). Just give instructions on the experiments.
Be consistent in your methodology and experiment design, for example, if you will use black-box LLM APIs such as GPT and Claude for your experiments, then you shouldn't propose any experiments that require white-box model weights or data access and you should edit them accordingly to follow the black-box assumptions.
Below are a few examples of how the full experiment plans should look like:
{examples}

Now please write down your experiment plan in JSON format (keys should be the section names, just like the above examples). Make sure to be as detailed as possible so that a student can directly follow the plan to implement the project.
\end{Verbatim}
\end{llmprompt}

\subsubsection{The AI Scientist}
We use The AI Scientist's \citep{lu2026towards} released ideation prompts with the AI-specific wording removed: ``ambitious AI PhD student'' becomes ``ambitious PhD student'' in the system prompt, and the first-round prompt introduces the task with ``You are given the following topic to work with:'' in place of the released experiment-template description and drops ``with the code provided'', since LiveIdeaBench supplies a keyword rather than an experiment template (the \texttt{<experiment.py>} block is then empty). The reflection prompt is as released.

\begin{llmprompt}{The AI Scientist: System Prompt}
\begin{Verbatim}[breaklines=true, breaksymbolleft=]
You are an ambitious PhD student who is looking to publish a paper that will contribute significantly to the field.
\end{Verbatim}
\end{llmprompt}

\begin{llmprompt}{The AI Scientist: First-Round Prompt}
\begin{Verbatim}[breaklines=true, breaksymbolleft=]
You are given the following topic to work with: {task_description}
<experiment.py>
{code}
</experiment.py>

Here are the ideas that you have already generated:

'''
{prev_ideas_string}
'''

Come up with the next impactful and creative idea for research experiments and directions you can feasibly investigate.
Note that you will not have access to any additional resources or datasets.
Make sure any idea is not overfit the specific training dataset or model, and has wider significance.

Respond in the following format:

THOUGHT:
<THOUGHT>

NEW IDEA JSON:
```json
<JSON>
```

In <THOUGHT>, first briefly discuss your intuitions and motivations for the idea. Detail your high-level plan, necessary design choices and ideal outcomes of the experiments. Justify how the idea is different from the existing ones.

In <JSON>, provide the new idea in JSON format with the following fields:
- "Name": A shortened descriptor of the idea. Lowercase, no spaces, underscores allowed.
- "Title": A title for the idea, will be used for the report writing.
- "Experiment": An outline of the implementation. E.g. which functions need to be added or modified, how results will be obtained, ...
- "Interestingness": A rating from 1 to 10 (lowest to highest).
- "Feasibility": A rating from 1 to 10 (lowest to highest).
- "Novelty": A rating from 1 to 10 (lowest to highest).

Be cautious and realistic on your ratings.
This JSON will be automatically parsed, so ensure the format is precise.
You will have {num_reflections} rounds to iterate on the idea, but do not need to use them all.
\end{Verbatim}
\end{llmprompt}

\begin{llmprompt}{The AI Scientist: Reflection-Round Prompt}
\begin{Verbatim}[breaklines=true, breaksymbolleft=]
Round {current_round}/{num_reflections}.
In your thoughts, first carefully consider the quality, novelty, and feasibility of the idea you just created.
Include any other factors that you think are important in evaluating the idea.
Ensure the idea is clear and concise, and the JSON is the correct format.
Do not make things overly complicated.
In the next attempt, try and refine and improve your idea.
Stick to the spirit of the original idea unless there are glaring issues.

Respond in the same format as before:
THOUGHT:
<THOUGHT>

NEW IDEA JSON:
```json
<JSON>
```

If there is nothing to improve, simply repeat the previous JSON EXACTLY after the thought and include "I am done" at the end of the thoughts but before the JSON.
ONLY INCLUDE "I am done" IF YOU ARE MAKING NO MORE CHANGES.
\end{Verbatim}
\end{llmprompt}

\subsubsection{The AI Scientist-v2}
We use The AI Scientist-v2's \citep{lu2026towards} released ideation prompts with two words removed from the system prompt: ``experienced AI researcher'' becomes ``experienced researcher'' and ``top ML conferences'' becomes ``top conferences''. The first-round and reflection prompts and the Semantic Scholar search tool are as released; the workshop-description slot holds the task description.

\begin{llmprompt}{The AI Scientist-v2: System Prompt}
\begin{Verbatim}[breaklines=true, breaksymbolleft=]
You are an experienced researcher who aims to propose high-impact research ideas resembling exciting grant proposals. Feel free to propose any novel ideas or experiments; make sure they are novel. Be very creative and think out of the box. Each proposal should stem from a simple and elegant question, observation, or hypothesis about the topic. For example, they could involve very interesting and simple interventions or investigations that explore new possibilities or challenge existing assumptions. Clearly clarify how the proposal distinguishes from the existing literature.

Ensure that the proposal does not require resources beyond what an academic lab could afford. These proposals should lead to papers that are publishable at top conferences.

You have access to the following tools:

{tool_descriptions}

Respond in the following format:

ACTION:
<The action to take, exactly one of {tool_names_str}>

ARGUMENTS:
<If ACTION is "SearchSemanticScholar", provide the search query as {"query": "your search query"}. If ACTION is "FinalizeIdea", provide the idea details as {"idea": { ... }} with the IDEA JSON specified below.>

If you choose to finalize your idea, provide the IDEA JSON in the arguments:

IDEA JSON:
```json
{
  "idea": {
    "Name": "...",
    "Title": "...",
    "Short Hypothesis": "...",
    "Related Work": "...",
    "Abstract": "...",
    "Experiments": "...",
    "Risk Factors and Limitations": "..."
  }
}
```

Ensure the JSON is properly formatted for automatic parsing.

Note: You should perform at least one literature search before finalizing your idea to ensure it is well-informed by existing research.
\end{Verbatim}
\end{llmprompt}

\begin{llmprompt}{The AI Scientist-v2: First-Round Prompt}
\begin{Verbatim}[breaklines=true, breaksymbolleft=]
{task_description}

Here are the proposals that you have already generated:

'''
{prev_ideas_string}
'''

Begin by generating an interestingly new high-level research proposal that differs from what you have previously proposed.
\end{Verbatim}
\end{llmprompt}

\begin{llmprompt}{The AI Scientist-v2: Reflection-Round Prompt}
\begin{Verbatim}[breaklines=true, breaksymbolleft=]
Round {current_round}/{num_reflections}.

In your thoughts, first carefully consider the quality, novelty, and feasibility of the proposal you just created.
Include any other factors that you think are important in evaluating the proposal.
Ensure the proposal is clear and concise, and the JSON is in the correct format.
Do not make things overly complicated.
In the next attempt, try to refine and improve your proposal.
Stick to the spirit of the original idea unless there are glaring issues.

If you have new information from tools, such as literature search results, incorporate them into your reflection and refine your proposal accordingly.

Results from your last action (if any):

{last_tool_results}
\end{Verbatim}
\end{llmprompt}

The system prompt substitutes the following text for
\texttt{\{tool\_descriptions\}}.

\begin{llmprompt}{The AI Scientist-v2: Tool Descriptions}
\begin{Verbatim}[breaklines=true, breaksymbolleft=]
- **SearchSemanticScholar**: Search for relevant literature using Semantic Scholar. Provide a search query to find relevant papers.

- **FinalizeIdea**: Finalize your idea by providing the idea details.

The IDEA JSON should include the following fields:
- "Name": A short descriptor of the idea. Lowercase, no spaces, underscores allowed.
- "Title": A catchy and informative title for the proposal.
- "Short Hypothesis": A concise statement of the main hypothesis or research question. Clarify the need for this specific direction, ensure this is the best setting to investigate this idea, and there are not obvious other simpler ways to answer the question.
- "Related Work": A brief discussion of the most relevant related work and how the proposal clearly distinguishes from it, and is not a trivial extension.
- "Abstract": An abstract that summarizes the proposal in conference format (approximately 250 words).
- "Experiments": A list of experiments that would be conducted to validate the proposal. Ensure these are simple and feasible. Be specific in exactly how you would test the hypothesis, and detail precise algorithmic changes. Include the evaluation metrics you would use.
- "Risk Factors and Limitations": A list of potential risks and limitations of the proposal.
\end{Verbatim}
\end{llmprompt}

The \texttt{\{tool\_names\_str\}} field is substituted as follows.

\begin{llmprompt}{The AI Scientist-v2: Tool Names}
\begin{Verbatim}[breaklines=true, breaksymbolleft=]
"SearchSemanticScholar", "FinalizeIdea"
\end{Verbatim}
\end{llmprompt}

\subsubsection{CAR}
\label{app:prompts_car}
The Idea Agent receives the system prompt below, rendered
for the full CAR arm. \texttt{\{node\_types\}} is filled
from the graph at runtime. The prompt gives short descriptions
of the operators. Their tool schemas separately specify
arguments and detailed instructions.

Freeform replaces the structured operators with
\texttt{creative\_reason}, which records natural-language
reasoning. Direct generates a proposal without the iterative
creative loop but retains the creativity-process instructions.
Floor also removes those instructions.
The leave-one-operator-out arms remove the corresponding
tool and its prompt entries. The no-operator-validation arm
changes the relevant validation instructions and checks,
while KG relation-label shuffle changes the graph's relation
labels. These variants require their corresponding prompt
and tool definitions to reproduce.

\begin{llmprompt}{CAR: Idea Agent System Prompt}
\begin{Verbatim}[breaklines=true, breaksymbolleft=]
You are an `Idea Agent` who aims to propose high-impact research ideas resembling exciting grant proposals using a given knowledge graph (your conceptual space) and extending it using your own knowledge / reasoning through various creativity processes. Combine the strengths of grounded knowledge from the graph, your existing knowledge and your creative reasoning to innovate beyond existing ideas. Make sure your ideas are novel. Be very creative and think out of the box. Clearly clarify how the proposal distinguishes from the existing literature.

AVAILABLE NODE TYPES:
The graph contains the following node types: {node_types}

AVAILABLE TOOLS:
1. Knowledge Graph Exploration:
    - `ask_knowledge_graph_exploration_agent`: Ask the `Knowledge Graph Exploration Agent` to retrieve graph-grounded information in different ways: queries without entity mentions, queries with entity mentions, and multi-entity or complex queries. Do not ask it to invent ideas or perform creative synthesis. You have at most 1 new KG exploration queries per idea (repeated cached questions do not count against this limit). Its answer lists relevant nodes as `[index] name (type)` (an identifier) along with its reasoning for why each node is relevant, NOT the node's full summary.
    - `get_entity_details`: Read the full summary (and any other metadata) of a node by its `[index]`. Use this to read the full text of the nodes you actually decide to build on — do not bulk-fetch every returned node because that is costly. Pass a single index, or a list of indices to fetch several entities at once.
    - `get_edges_by_entity`: List the real literature-derived edges (subject -[relation (description)]-> object) incident to a node `[index]` (or a list of indices), in both directions. Use it to understand the relational context around an entity. Excludes your proposed edges; not paginated.
    
2. Creative Thinking:
    - `bisociate`: Connect two distinct existing entities that are not directly linked, through one newly proposed intermediate bridge concept that emerges at their intersection
    - `analogize`: Transfer a relation you know onto a new pair: state a source triplet from your own knowledge (it need not exist in the graph, but must be a coherent, justified relation with a relation_description) and map that same relation onto a different target pair (neither target endpoint may be the source subject or object) while keeping at least one target entity grounded in the graph
    - `explore`: Extend exactly one existing entity to exactly one new entity, using a relation you may devise (justified by its relation_description) -- it need not be an existing graph relation
    - `elaborate`: Deepen one of your previously proposed concepts by making its description more specific, technical, and concretely implementable -- without introducing any new entity -- to develop one mechanism in depth instead of accumulating loosely-related concepts; you may optionally anchor it to an existing corpus entity it builds on or is realized through, via `grounding_edge`

3. Research Proposal Generation (Mandatory):
    - `case_resolved`: Signal that the proposal generation task is completed and provide the final proposal output using only the structured proposal fields: `name`, `title`, `short_hypothesis`, `related_work`, `abstract`, `experiments`, `risk_factors_and_limitations`, and `used_entities` (the `[index]` values of the existing KG entities the proposal is grounded in).

WORKFLOW:
1. Knowledge Acquisition:
    - Use `ask_knowledge_graph_exploration_agent` to gather knowledge in different ways: queries without entity mentions, queries with entity mentions, and multi-entity or complex queries. Do not use this tool to invent ideas or to perform creative synthesis. The `Knowledge Graph Exploration Agent` retrieves nodes only for grounding your creative thinking. The exploration answer gives you `[index] name (type) — short reasoning for selection` per node.
    - Use `get_entity_details` and `get_edges_by_entity` to get detailed information about entity node(s) and the real edges that you decide to build on. Do not fetch details for every returned node.
2. Creative Thinking:
    - At each creative step, choose one of the 4 following creative actions to extend the knowledge graph (along with detailed reasoning for your choice of action and how it contributes to scientific creativity):
        1. Combinational creativity (bisociation): Use `bisociate` to connect two distinct existing entities that are not directly linked, through one newly proposed intermediate bridge concept that emerges at their intersection
        2. Combinational creativity (analogical reasoning): Use `analogize` to transfer a relation you know onto a new pair: state a SOURCE triplet from your own knowledge (it need not exist in the graph, but must be a coherent, justified relation with a relation_description) and map that same relation onto a DIFFERENT target pair (neither target endpoint may be the source subject or object) while keeping at least one target entity grounded in the graph
        3. Exploratory creativity: Use `explore` to extend exactly one existing entity to exactly one new entity, using a relation you may devise (justified by its relation_description) -- it need not be an existing graph relation
        4. Elaboration: Use `elaborate` to deepen ONE of your previously proposed concepts by making its description more specific, technical, and concretely implementable -- without introducing any new entity -- to develop one mechanism in depth instead of accumulating loosely-related concepts; you may optionally anchor it to an existing corpus entity it builds on or is realized through, via `grounding_edge`
    - You may reuse the same tool multiple times. These tools are based on creativity processes: bisociation (Koestler, 1964), analogical reasoning (Gentner, 1983), exploratory creativity (Boden, 2004) and elaboration (Guilford, 1967).
    - Perform as many meaningful and non-trivial creative steps as needed to construct a research proposal with significant creative depth.
    - You MUST perform at least 3 successful creative steps before concluding.
    - ONE STEP AT A TIME: Do NOT execute multiple creative actions in a single turn. You must wait for the result of the current action (which modifies the knowledge graph) before deciding the next step. This ensures each step is grounded in the latest graph state.
    - ANCHOR ON A SPECIFIC ENTITY, NOT A PAPER: every creative action must operate on non-paper entities (concept, method, result, claim, problem, metric, etc.) — paper nodes are for provenance/scaffolding only.
    - Triplet/edge format: each edge is {"subject": <name>, "object": <name>, "relation": <short phrase>, "relation_description": <why this specific edge holds>}. When an endpoint is a NEW node, also include its type/description ("object_type"/"object_description" or "subject_type"/"subject_description").
    - If a creative step fails validation, you will receive an error message and should try again with a different variation of the same action or a different creative action altogether.
    - Continue until you have constructed a significantly novel and coherent scientific mechanism through your steps.
3. Research Proposal Generation:
    - Once your SUCCESSFUL creative actions can synthesize a novel and useful proposal, use `case_resolved` to output your final research proposal using only the structured proposal fields: `name`, `title`, `short_hypothesis`, `related_work`, `abstract`, `experiments`, `risk_factors_and_limitations`, and `used_entities` (the `[index]` values of the existing KG entities the proposal is grounded in).
    - Ensure a novel and deep synthesis of your proposed concepts, centered on one novel core mechanism. Be selective: integrate only the concepts that directly serve the central hypothesis, rather than blindly combining all your proposed concepts into a kitchen sink of loosely related concepts.
    - Ensure the proposal is clear and the JSON is in the correct format.

GENERAL GUIDELINES:
- Keep the emerging proposal focused around one central hypothesis. Prefer a single, clean novel mechanism synthesized in a novel way over a broad collection of loosely related mechanisms.
- Avoid trivial combinations. Do not simply combine existing ideas or methods directly from the knowledge graph. You must propose a novel direction that goes beyond a simple mix of components.
- Make sure your proposal is grounded in the creative steps you have taken.
- Ensure ideas are novel yet feasible.
- These proposals should lead to papers that are publishable at top conferences.
- Include mathematical formulas when relevant. Note that any math formula should be as complete as possible.
- Make sure any proposal does not overfit the specific training dataset or model, and has wider significance.
- Note that you will not have access to any additional external datasets.
\end{Verbatim}
\end{llmprompt}

Each idea starts from the per-idea query below. The research task is the LiveIdeaBench keyword verbatim, the paper list is the frozen corpus for that keyword, and the last two fields carry the anti-redundancy memory: the exploration questions asked for earlier ideas in the run (questions only, never their answers) and the earlier proposals themselves.
\begin{llmprompt}{CAR: Idea Agent Per-Idea Query}
\begin{Verbatim}[breaklines=true, breaksymbolleft=]
{research_task}

The knowledge graph contains scientific entities and relations from the following papers:
{papers}

Here are the knowledge graph exploration questions you have already asked for previous proposals:

'''
{previous_kg_queries}
'''

Prefer exploring different topics and concepts than these - avoid repeating the same exploration routes.

Here are the proposals that you have already generated:

'''
{previous_proposals}
'''

Begin by generating an interestingly new high-level research proposal that differs from what you have previously proposed.
\end{Verbatim}
\end{llmprompt}

Graph queries are delegated to a separate exploration agent
adapted from \citet{polonuer2026autonomous}.
It selects and justifies nodes using search snippets and
neighborhood relations, rather than generating proposals.
The \texttt{\{node\_types\}} and
\texttt{\{relation\_types\}} placeholders are filled from
the graph at runtime.

\begin{llmprompt}{CAR: Knowledge Graph Exploration Agent System Prompt}
\begin{Verbatim}[breaklines=true, breaksymbolleft=]
You are a `Knowledge Graph Exploration Agent` specialized in exploring a knowledge graph to find specific entities and metadata that answer complex questions. The graph structure, entity types and relations vary by domain, but the exploration strategy remains consistent.

AVAILABLE NODE TYPES:
The graph contains the following node types: {node_types}
Between the nodes, the possible relation types are: {relation_types}

AVAILABLE TOOLS:
1. Knowledge Graph Exploration:
    - `search_in_graph`: Use this for initial broad searches across the entire graph to identify relevant entities. The search uses BM25, which works well for keyword-based retrieval.
    - `search_in_neighborhood`: Use this to explore specific nodes found through initial searches. Shows relation types between the reference node and its neighbors with directional arrows.
    - Nodes shown as `(proposed TYPE)` were proposed by the `Idea Agent` during earlier creative steps; treat them as speculative working memory, not literature evidence.

2. Answer Reporting (Mandatory):
    - `add_to_answer`: Use this to collect relevant entities with justifications.
    - `finish`: Use this when you have found all relevant nodes or exhausted useful exploration paths.

WORKFLOW:
1. You will receive a question about the knowledge graph.
2. Search in graph for the entities mentioned in the question.
3. Add the relevant results to the answer.
4. See their neighbors to see if they contain more information.
5. Add the relevant neighbors.
6. Call finish.

EXPLORATION STRATEGY:
Your approach should adapt based on the query structure:
1. Queries without explicit entity mentions
    - When the query does not explicitly mention specific entities (e.g., product names, paper titles, gene names, author names, etc.), use a broad search strategy to provide the user with many options:
        a. Use `search_in_graph` with the full question as the query and size=30 to cast a wide net
        b. Review all 30 results and select approximately 15 of the most suitable entities to add to the answer (aim for roughly half of the search results)
        c. Add the selected entities to the answer with clear reasoning for each
    - **Important**: The goal is to provide users with multiple relevant options. When the query is descriptive and doesn't name specific entities, you should return a substantial number of results (around 15 from a search of size 30). This ensures users have many options to choose from. Only exclude results that are clearly irrelevant to the query.
    - This strategy works well when:
        - The query is descriptive but doesn't name specific entities
        - You want to provide multiple options to the user (which is often the case)
        - The graph's search function (BM25) can effectively match keywords from the query

2. Queries with explicit entity mentions
    - When the query explicitly mentions specific entities (e.g., "product X", "paper Y", "gene Z", "author W"), use a targeted exploration strategy:
        a. **Entity disambiguation**: First, search for the mentioned entities using `search_in_graph` with the entity name
        b. **Neighborhood exploration**: Once you've identified the relevant entity nodes, use `search_in_neighborhood` to explore their connections
        c. **Filtered search**: Use the `query` parameter in neighborhood searches to filter results by keywords from the original question
    - This strategy works well when:
        - The query mentions specific entities that likely exist in the graph
        - You need to explore relationships around known entities
        - The query requires multi-hop reasoning

3. Multi-entity or complex queries
    - For queries that involve multiple entities or require combining information:
        a. Start by disambiguating all mentioned entities
        b. Explore neighborhoods of key entities with relevant filters
        c. Combine information from multiple exploration paths

EXAMPLES:
1. Simple broad search (Strategy 1)
    - **Query**: "Find products suitable for outdoor camping"
    - **Approach**: Since no specific products are mentioned, use `search_in_graph(query="Find products suitable for outdoor camping", size=30)`. This will return 30 results. Then, review all 30 results and select approximately 15 of the most suitable products to add to the answer. This gives the user many options to choose from.

2. Entity-specific query (Strategy 2)
    - **Query**: "What are some winter-themed accessories from the BrandX company?"
    - **Approach**: 
        a. First, search for the brand/company: `search_in_graph("BrandX company")`
        b. Then explore its neighborhood: `search_in_neighborhood(node_index=<found_brand_index>, query="winter-themed accessories")`

3. Multi-hop reasoning (Strategy 3)
    - **Query**: "Can you find other publications from the co-authors of the paper titled 'Machine Learning Applications in Healthcare' that relate to neural networks?"
    - **Approach**:
        a. Search for the paper: `search_in_graph("Machine Learning Applications in Healthcare")`
        b. Find co-authors: `search_in_neighborhood(node_index=<paper_index>, node_type=author)`
        c. For each author, search their papers: `search_in_neighborhood(node_index=<author_index>, query="neural networks", node_type=paper)`

4. Multiple constraints (Strategy 4)
    - **Query**: "What medications interact synergistically with DrugX and are also used to treat DiseaseY?"
    - **Approach**:
        a. Find DrugX: `search_in_graph("DrugX")`
        b. Find DiseaseY: `search_in_graph("DiseaseY")`
        c. Explore neighborhoods: `search_in_neighborhood(node_index=<disease_index>, node_type=drug)` and `search_in_neighborhood(node_index=<drugx_index>, node_type=drug)`
        d. Add the drugs to the answer

GENERAL GUIDELINES:
- **Provide multiple options when appropriate**: For queries without explicit entity mentions, aim to give users many relevant options (typically 10-20 entities from a search of size 30)
- **Start broad, then narrow**: Begin with global searches, then focus on specific neighborhoods
- **Use filters strategically**: Apply `node_type` and `edge_type` filters to reduce noise and focus exploration
- **Combine multiple strategies**: Complex queries may require mixing broad searches and neighborhood exploration
- **Balance relevance and coverage**: When selecting entities to add to answers, prioritize relevance but also aim for good coverage when the query allows for multiple valid answers
- **Provide reasoning**: Always include clear reasoning when adding entities to answers
- **Adapt to graph characteristics**: Some graphs may benefit more from broad searches (e.g., when BM25 works well), while others may require more targeted exploration
- **Be informative**: Provide informative and accurate answers grounded in the search snippets and neighborhood relations you retrieve
\end{Verbatim}
\end{llmprompt}

\subsection{LiveIdeaBench}
\label{app:prompts_liveideabenchoriginal}

\subsubsection{100-word idea conversion}
We adapt the benchmark's idea-generation instruction to rewrite
an existing proposal, targeting a length of $100$ words.

\begin{llmprompt}{Idea Conversion to 100-Word Abstract Prompt}
\begin{Verbatim}[breaklines=true, breaksymbolleft=]
I'll be submitting your next response to a "Good Scientific Idea" expert review panel. If they consider your idea to be a good one, you'll receive a reward. Your assigned keyword is: "{keyword}". You may provide background information. The idea MUST be concisely expressed within 100 words total (including any background information). (Note: good scientific ideas should be original (novel contribution), feasible (technically implementable), clearly articulated, and address meaningful problems in the field.)."

Rewrite your research proposal into a single, self-contained "Good Scientific Idea".

Your research proposal is:
{idea}
\end{Verbatim}
\end{llmprompt}

\subsubsection{Scoring}
We use LiveIdeaBench's scoring prompts \citep{ruan2026evaluating}.
The originality, feasibility and clarity critic receives the first
prompt as its system message and the second as its user message.
The \texttt{\{idea\}} field contains the idea text used for evaluation.
Fluency is judged on every unordered pair of available ideas from
the same method and keyword, with answers mapped to
A${}=10$, B${}=7$, C${}=4$ and D${}=1$.

\begin{llmprompt}{{Originality, Feasibility and Clarity Critic Prompt}}
\begin{Verbatim}[breaklines=true, breaksymbolleft=]
You are an extremely demanding scientific reviewer with the highest critical standards, like those at Nature or Science. When evaluating scientific ideas, you will assess them on three key dimensions:

1. originality: Novel contribution to unexplored areas or innovative approaches to existing problems
2. feasibility: Technical implementation and practicality
3. clarity: How well-articulated and easy to understand the idea is

Your response should consist of two parts: a text analysis followed by a JSON score block.

First, provide your brief analysis (less than 100 words) of the idea. Then, for each dimension, provide a score from 1 to 10 where 1-3 = poor, 4-6 = average, 7-10 = excellent.

For example:
```json
{
    "originality": <score_1_to_10>,
    "feasibility": <score_1_to_10>,
    "clarity": <score_1_to_10>
}```
\end{Verbatim}
\end{llmprompt}

\begin{llmprompt}{{Originality, Feasibility and Clarity User Prompt}}
\begin{Verbatim}[breaklines=true, breaksymbolleft=]
Please evaluate the following scientific idea:

{idea}
\end{Verbatim}
\end{llmprompt}

\begin{llmprompt}{Fluency Critic Prompt}
\begin{Verbatim}[breaklines=true, breaksymbolleft=]
Here are two ideas submitted to "Good Scientific Ideas" Competition, which both relate to "{keyword}":

# The first idea

{A}

# The second idea

{B}


# Question

Evaluate the similarity between these two ideas that both relate to "{keyword}". Please choose the best answer:

A. Completely different ideas addressing different problems, despite relating to the same keyword.
B. Different ideas but addressing similar problems.
C. Similar ideas addressing similar or identical problems.
D. Academically identical ideas with the same core approach and problem statement.

ONLY ANSWER A/B/C/D, DO NOT EXPLAIN
\end{Verbatim}
\end{llmprompt}

\subsection{LiveIdeaBench (methodological novelty judge)}
\label{app:prompts_overlapjudge}
\subsubsection{250-word idea conversion}
\begin{llmprompt}{Idea Conversion to 250-Word Abstract Prompt}
\begin{Verbatim}[breaklines=true, breaksymbolleft=]
Convert the following research idea into an abstract that summarizes the proposal in conference format (approximately 250 words).

Return only the abstract paragraph text. Do not include a heading, label, title, markdown, bullet list, or JSON.

Source title:
{title}

Source full idea:
{idea}

Abstract:
\end{Verbatim}
\end{llmprompt}

\subsubsection{Literature retrieval}
Retrieval adapts the iterative Semantic Scholar loop of
\citet{si2025can}. It starts with a keyword query in round $0$
and continues for up to ten further rounds, stopping early
once the paper bank contains at least $120$ papers.
For subsequent queries, the model sees up to ten papers with
the highest relevance scores and its previous queries, then
issues one new \texttt{KeywordQuery}, \texttt{PaperQuery} or
\texttt{GetReferences} call. If no papers have been collected,
the initial query prompt is reused.
In both retrieval and relevance scoring, \texttt{\{idea\}}
contains the idea's title followed by its approximately
$250$-word abstract. Each grounding paper is listed by
\texttt{paperId} and title.
The prompts encourage short queries in established terminology
rather than terms coined by the proposed idea.
\begin{llmprompt}{{Retrieval Prompt, Round 0}}
\begin{Verbatim}[breaklines=true, breaksymbolleft=]
You are a researcher doing a literature review to find the most relevant prior work to judge whether the following research idea is novel:
{idea}

You should propose some keywords for querying the Semantic Scholar API to find the most relevant papers to the research idea. Use the standard terminology a researcher in this field would already use - search engines can only match wording that exists in published papers, so do not use the idea's own invented compound terms.

Formulate your query as: KeywordQuery("query"). Just give me one query, and the query can be a concatenation of multiple keywords (just put a space between every word); but please keep it short (3-6 words; an over-long, over-specific query returns nothing) and try to cover the most important aspects.

Your query (just return the query itself with no additional text):
\end{Verbatim}
\end{llmprompt}
\begin{llmprompt}{{Retrieval Prompt, Rounds 1--10}}
\begin{Verbatim}[breaklines=true, breaksymbolleft=]
You are a researcher doing a literature review to find the most relevant prior work to judge whether the following research idea is novel:
{idea}

You are allowed to use the following functions for making queries:
(1) KeywordQuery("keyword"): find most relevant papers to the given keyword (the keyword should be short and in established terminology, otherwise the search engine will return nothing; it is ok to combine a few short keywords with spaces, such as "language model reasoning").
(2) PaperQuery("paperId"): find the most relevant papers to the given paper (as specified by the paperId).
(3) GetReferences("paperId"): get the list of papers referenced in the given paper (as specified by the paperId).

Right now you have already collected the following relevant papers:
{grounding}

You can formulate new search queries based on these papers. And you have already asked the following queries:
{queries}

Please formulate a new query to expand our paper collection with more diverse and relevant papers (you can do so by diversifying the types of queries and minimizing the overlap with previous queries; e.g., try to include all three types of queries if possible). Directly give me your new query without any explanation or additional text, just the query itself:
\end{Verbatim}
\end{llmprompt}

\subsubsection{Literature ranking}
Retrieved papers must have a paper identifier and an abstract
longer than $20$ words to enter the paper bank.
Newly added papers are scored for their usefulness as prior
art using a prompt adapted from \citet{si2025can}, with criteria
focused on relevance for judging novelty.
Each paper is listed by \texttt{paperId}, title and full abstract.
For the final evidence set, papers are ranked by this relevance
score, with ties broken by SPECTER2 similarity
\citep{singh2023scirepeval}. The top ten papers are selected
for judging.
\begin{llmprompt}{Prior-Art Relevance Scoring Prompt}
\begin{Verbatim}[breaklines=true, breaksymbolleft=]
Research idea being judged for novelty:
{idea}

You are a helpful literature review assistant whose job is to read the below set of papers and score each paper on how useful it is for judging whether the research idea above is novel. The criteria for scoring are:
(1) The paper is directly relevant to the idea's core methodology or the specific problem it addresses -- it would be cited as related work. A paper on the same broad topic but a different mechanism scores low.
(2) The paper proposes a concrete method, rather than being a survey, position, or analysis paper.
(3) The paper is close enough that a reviewer would want to see it when asking "has this been done before?".

The papers are:
{papers}

Please score each paper from 1 to 10. Write the response in JSON format with "paperId": score as the key and value for each paper.
\end{Verbatim}
\end{llmprompt}

\subsubsection{Scoring}
The judge prompt adapts the stratified novelty judge of
\citet{shen2026unlocking} from its biomedical setting to the
research task specified by the LiveIdeaBench keyword
(\texttt{\{task\_description\}}, passed verbatim).
Retrieved papers are numbered from $0$, each with its title
and abstract. All three panel judges receive the same input.
Although the prompt requests both \texttt{methodology\_overlap}
and \texttt{novelty\_score}, we compute the reported MN as
$10$ minus the mean of the three
\texttt{methodology\_overlap} values.
Ideas without eligible retrieved evidence or a complete
three-judge panel are excluded.
\begin{llmprompt}{Methodological Novelty Judge System Prompt}
\begin{Verbatim}[breaklines=true, breaksymbolleft=]
You are a senior top-tier conference and journal reviewer judging whether the specific methodology of a proposed research idea has already been applied to its target research task, given the most relevant retrieved papers.
\end{Verbatim}
\end{llmprompt}
\begin{llmprompt}{Methodological Novelty Judge User Prompt}
\begin{Verbatim}[breaklines=true, breaksymbolleft=]
Proposed idea's title: {title}
Proposed idea's abstract: {abstract}
Research area: {task_description}

Papers found that may describe similar work:
{prior_block}

First, identify the core methodology/mechanism of the proposed solution. Then determine if the SPECIFIC METHODOLOGY has already been applied to THIS SPECIFIC research task.

CRITICAL: The novelty we care about is whether this methodology has been applied to THIS task before - NOT whether the methodology exists in general. Finding papers that use the same method on other tasks or domains (climate, finance, etc.) does NOT reduce novelty - that is exactly what cross-domain transfer is about. Retrieval always returns topically-related papers, so judge whether they actually match the core method, not whether related work exists.

Rate methodology overlap (0-10) - BE STRICT:

0-1: REVOLUTIONARY - This exact methodology has NEVER been explored for this task. Zero papers use this approach or related variants; a paradigm-shifting first application.

2-3: HIGHLY NOVEL - The specific methodology has not been applied, but some distant conceptual relatives exist. Retrieved papers address the task with completely different approaches; any methodological similarities are superficial.

4-6: MODERATE NOVELTY - Related methodologies exist but this specific approach has not been used. Same family of methods but different specific techniques/formulations; the core algorithmic innovation has not been applied to this task.

7-8: LIMITED NOVELTY - Very similar methodologies already exist for this task. Most key components are already present in existing work; a minor variation or incremental extension.

9-10: NOT NOVEL - This exact methodology has been applied to this exact task. Retrieved papers describe essentially the same approach; direct prior art exists.

Return ONLY valid JSON:
{"methodology_overlap": N, "novelty_score": N, "assessment": "brief explanation - be explicit about whether the SPECIFIC methodology has been applied to this task or not"}

Where novelty_score = 10 - methodology_overlap (higher = more novel, range 0-10)
\end{Verbatim}
\end{llmprompt}

\end{document}